\documentclass[10pt,twocolumn,letterpaper]{article}

\usepackage[pagenumbers]{cvpr} 

\usepackage{graphicx}
\usepackage{amsmath}
\usepackage{amssymb}
\usepackage{booktabs}

\usepackage{multirow}
\usepackage{bm}
\usepackage{bbm}
\usepackage{wrapfig}
\usepackage{caption}
\usepackage[table, dvipsnames]{xcolor}

\newcommand{\cmark}{\checkmark}

\definecolor{gg}{gray}{0.9}

\usepackage[pagebackref,breaklinks,colorlinks]{hyperref}

\usepackage[capitalize]{cleveref}
\crefname{section}{Sec.}{Secs.}
\Crefname{section}{Section}{Sections}
\Crefname{table}{Table}{Tables}
\crefname{table}{Tab.}{Tabs.}

\begin{document}


\title{Self-Supervised Visual Representation Learning via Residual Momentum}

\author{Trung X. Pham$^{1}$ \ \ Axi Niu$ ^2$ \ \ Zhang Kang$^1$ \ \ Sultan Rizky Madjid$^1$ \ \ Ji Woo Hong$^1$ \\
Daehyeok Kim$^1$ \ \ Joshua Tian Jin Tee$^1$ \ \ Chang D. Yoo$^1$ \\
$^1$Korea Advanced Institute of Science and Technology (KAIST) \\
$^2$Northwest Polytechnical University Xi'an \\
{\tt\small $^1$\{trungpx,zhangkang,suulkyy,jiwoohong93,kimshine,joshuateetj,cd\_yoo\}@kaist.ac.kr} \\
{\tt\small $^2$nax@mail.nwpu.edu.cn }
}

\maketitle

\begin{abstract}
    Self-supervised learning (SSL) approaches have shown promising capabilities in learning the representation from unlabeled data. Amongst them, momentum-based frameworks have attracted significant attention. Despite being a great success, these momentum-based SSL frameworks suffer from a large gap in representation between the online encoder (student) and the momentum encoder (teacher), which hinders performance on downstream tasks. This paper is the first to investigate and identify this invisible gap as a bottleneck that has been overlooked in the existing SSL frameworks, potentially preventing the models from learning good representation. To solve this problem, we propose ``residual momentum'' to directly reduce this gap to encourage the student to learn the representation as close to that of the teacher as possible, narrow the performance gap with the teacher, and significantly improve the existing SSL. Our method is straightforward, easy to implement, and can be easily plugged into other SSL frameworks. Extensive experimental results on numerous benchmark datasets and diverse network architectures have demonstrated the effectiveness of our method over the state-of-the-art contrastive learning baselines.

\end{abstract}

\vspace{-16pt}
\section{Introduction}
\vspace{-4pt}
\label{intro}
Self-supervised learning (SSL) has had great success in NLP \cite{brown2020language,devlinetal2019bert} 
in the past few years. Recently, it also became an emerging paradigm and critical research for computer vision, thanks to its unique advantages of not requiring any costly labeling process from humans as supervised learning frameworks do \cite{he2016deep,ren2015faster,pham2022lad,kim2020modality,9718302niu}. SSL has surpassed traditional supervised pretraining methods in learning representations for many downstream tasks such as classification, segmentation, and object detection \cite{wang2020DenseCL,he2020momentum,chen2021mocov3,caron2021emerging,grill2020bootstrap,madaan2022representational,zhang2022how,zhang2022dual,robinson2021contrastive}.
Without the ground-truth label, the core of most SSL methods lies in learning an encoder using augmentation-invariant representation \cite{bachman2019learning,he2020momentum,chen2020simple,caron2020unsupervised,grill2020bootstrap}. Amongst them, SSL frameworks based on exponential moving averages (EMA or momentum) have attracted much attention.
MoCo \cite{he2020momentum}, MoCo-v2 \cite{chen2020improved}, BYOL \cite{grill2020bootstrap}, DINO \cite{caron2021emerging}, ReSSL \cite{zheng2021ressl}, iBOT \cite{zhou2022image_ibot}, and the recent MoCo-v3 \cite{chen2021mocov3} are examples of the momentum-based frameworks
that achieved great success in self-supervised visual representation learning. These frameworks use EMA to construct two branches of a Siamese architecture where one branch is with momentum encoder (called ``teacher'' \cite{caron2021emerging} or ``target'' encoder \cite{grill2020bootstrap}) and the other branch without it (called ``student'' or online encoder \cite{grill2020bootstrap}).
\begin{figure}[!tbp]
  \vspace{-10pt}
  \centering
  \subfloat[CIFAR-100 \& ImageNet-1K] {\includegraphics[width=0.5\linewidth]{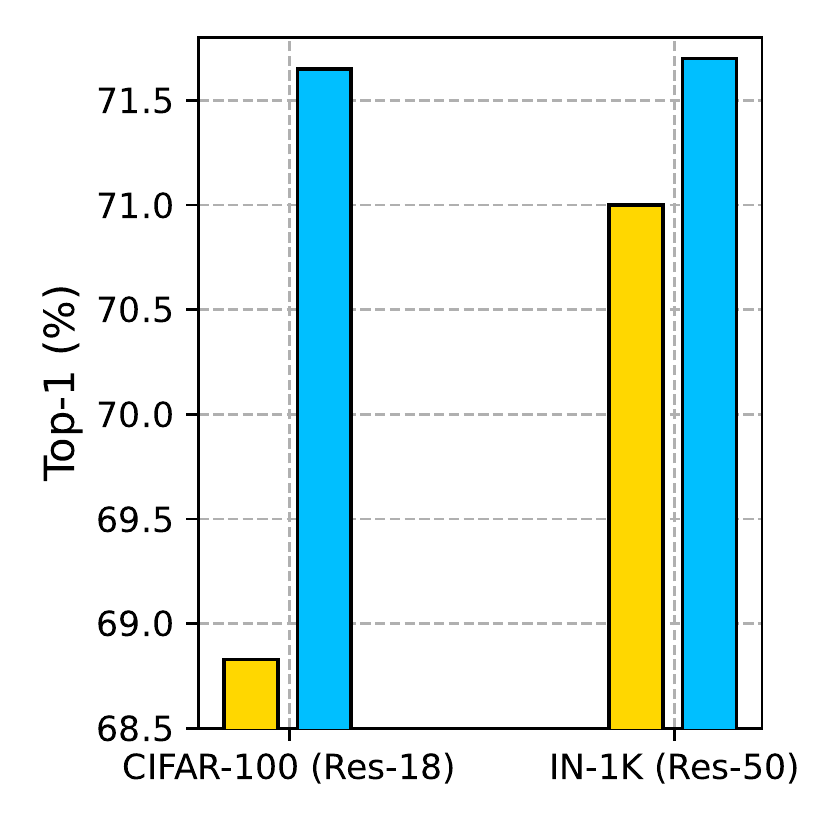}}
  \hfill
  \subfloat[CIFAR-10 \& ImageNet-100] {\includegraphics[width=0.5\linewidth]{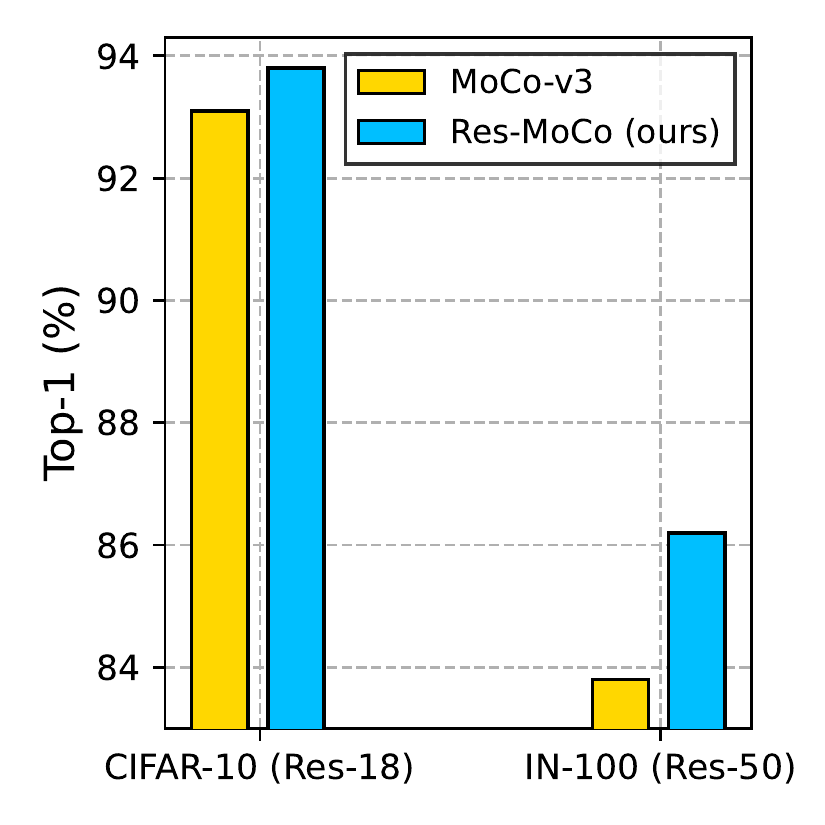}}
  \vspace{-10pt}
  \caption{Linear classification comparison on four datasets. We use ResNet-18 for CIFAR-10 and CIFAR-100; ResNet-50 for ImageNet-100 and ImageNet-1K. All models are trained for 1000 epochs except large ImageNet-1K we train with 200 epochs.}
  \label{fig:ema_whole_part}
  \vspace{-12pt}
\end{figure}
Specifically, MoCo \cite{he2020momentum} is a milestone that first applies EMA in SSL by introducing a slow-moving average network (momentum encoder) to maintain consistent representations of negative pairs to a large memory bank. Without negative samples, BYOL \cite{grill2020bootstrap} uses a moving average network to produce prediction targets to stabilize the bootstrap step and a simple cosine similarity loss to reduce the distance between the two distorted versions (positive pairs) of an image. Another work DINO \cite{caron2021emerging}, designs an SSL approach in the form of knowledge distillation to predict the output of the teacher encoder constructed by a momentum encoder as the target and minimize the distance between the student and teacher by using a standard cross-entropy loss. More recently, MoCo-v3 \cite{chen2021mocov3} employed the best practices in the area to achieve a more powerful framework. Those momentum-based approaches have greatly enriched the field of self-supervised learning; however, they only focus on reducing the distance between representations of two different augmented views (\textbf{inter-view}) while overlooking the impact of reducing the distance between representations of the teacher and student for the same augmented view (\textbf{intra-view}).
\begin{figure}[!tbp]
  \vspace{-12pt}
  \centering
  \includegraphics[width=1.0\linewidth]{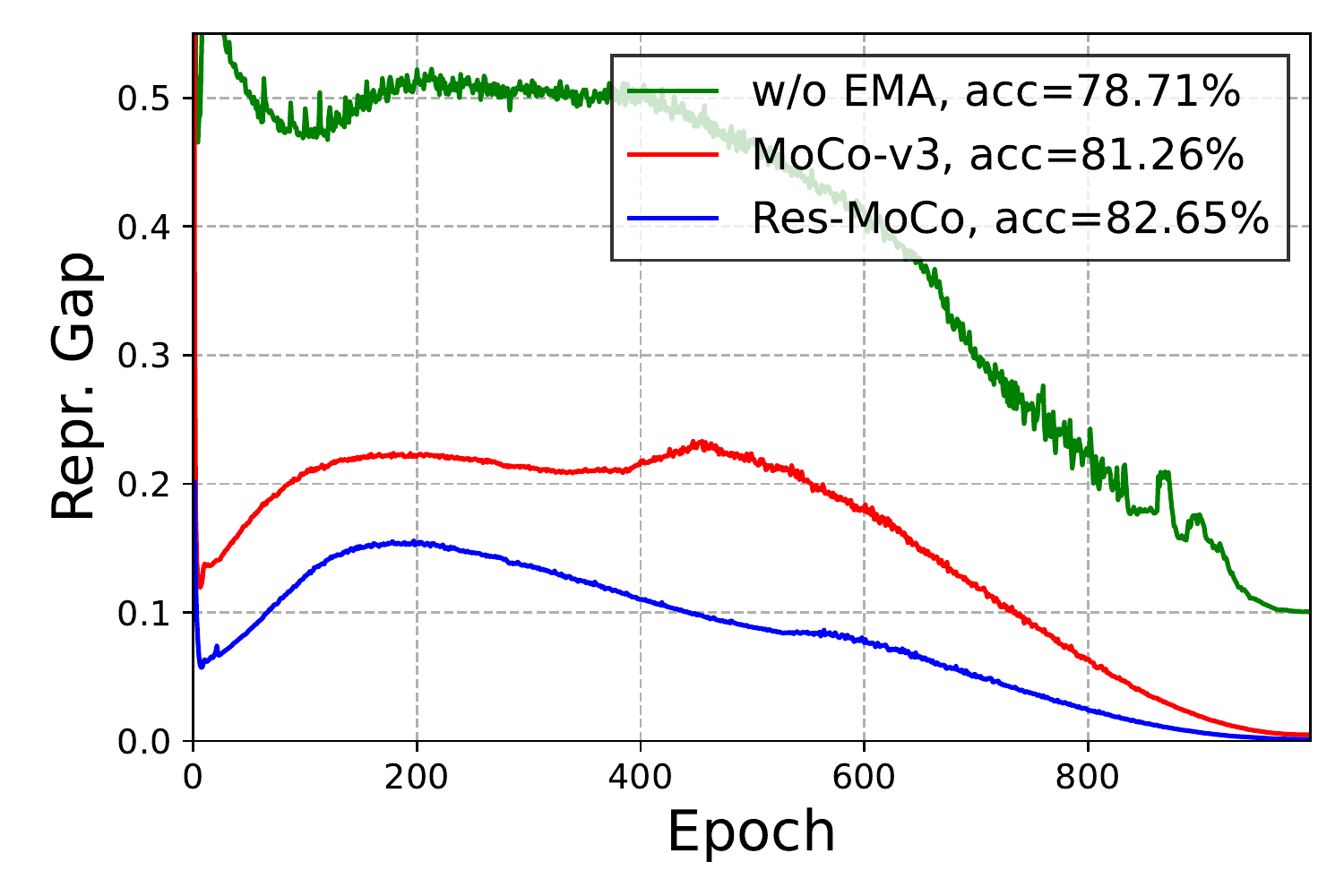} 
  \vspace{-13pt}
  \caption{Representation gap between the teacher and student of MoCo-v3, MoCo-v3 w/o EMA, and and Res-MoCo (MoCo-v3 w/ Intra-M) on ImageNet-100, measured by Eq. \ref{eq:cosine_loss}.}
  \vspace{-16pt}
  \label{fig:huge_representation_gap}
\end{figure}
We investigate the potential gap between the teacher and student from the exact same input augmentation (\textbf{intra-representation gap}) instead of the different augmentations as considered in the existing works (\textbf{inter-representation gap}) as illustrated in Fig. \ref{fig:architecture}. Interestingly, we find that a considerable \textit{intra-representation gap} 
is presented in the existing SSL as shown in Fig. \ref{fig:huge_representation_gap}. We show that such a considerable gap consequentially causes a large performance gap between the teacher and student, and it may potentially prevent the SSL models from learning better representation (Fig. \ref{fig:compare_gap_teacher_student_cifar100}).
To this end, we propose \textit{residual momentum} (we also refer to as \textit{intra-momentum} to differentiate with \textit{inter-momentum} in existing SSLs) to reduce the intra-representation gap systematically. In this way, the student model is trained to match the teacher's output closely, narrowing their performance gap and hence improving the performance of both models. Note that in EMA-based SSL, the teacher's weight is dynamically updated based on the student's weight \cite{grill2020bootstrap,caron2021emerging,chen2021mocov3}, so if the student improves, the teacher gets better and vice versa (Fig. \ref{fig:compare_gap_teacher_student_cifar100}). 
Overall, our contributions are summarized as follows:
\vspace{-2pt}
\begin{itemize}
    \item To the best of our knowledge, our work is the first attempt to investigate the potential intra-representational gap between the encoders of teacher and student in EMA-based SSLs. We identify this often ignored but crucial gap causing a huge performance gap between the teacher and student, potentially preventing the student from learning better representations.
    \vspace{-2pt}
    \item We propose Res-MoCo with \textit{intra-momentum} in the training process to directly reduce the intra-representation gap between teacher and student encoders in the SSL frameworks, so significantly narrowing their performance gap and boosting the performance of the student model.
    \vspace{-2pt}
    \item Extensive experimental results on different benchmark datasets and various network architectures demonstrate the effectiveness of the proposed method over the state-of-the-art SSL baselines.
\end{itemize}
\begin{figure*}[!htbp]
  \centering
  \includegraphics[width=0.85\linewidth]{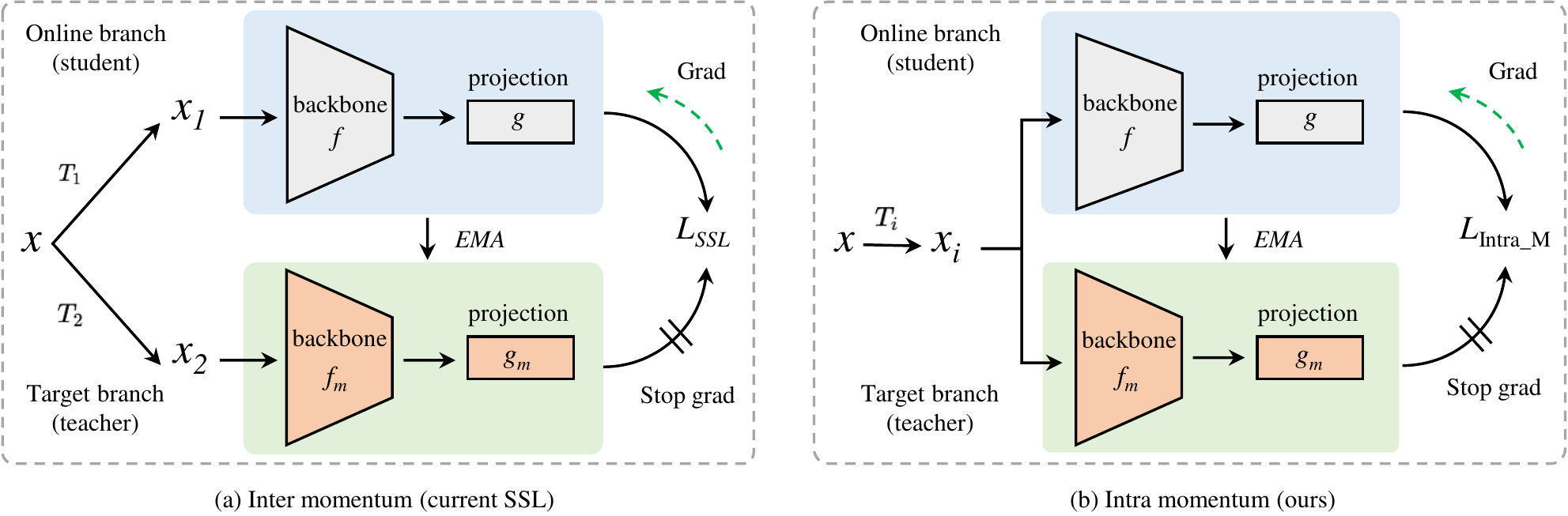}
  \vspace{-6pt}
  \caption{Compare the existing SSL frameworks that use Inter-momentum and the proposed method with Intra-momentum. $T_i$ with $i\in \{1,2\}$ is a random transformation to the given input image $x$. As the term suggests, existing SSL with Inter-M uses different inputs, \ie $x_1$ and $x_2$ for student and teacher, respectively. By contrast, a model with Intra-momentum uses exactly the same input $x_i$ with $i\in \{1,2\}$ for both the student and teacher model and uses a distance function loss $\mathcal{L}_{\text{Intra\_M}}$ to minimize the intra-representation gap.}
  \label{fig:architecture}
  \vspace{-14pt}
\end{figure*}

\section{Related Works}
\label{related_works}
\textbf{Momentum-Based Self-Supervised Learning.}
Exponential moving average (EMA or momentum) has been deeply studied for smoothing the original sequence signal \cite{pring2002technical,klinker2011exponential,appel2005technical},
optimization \cite{kingma2014adam,sutskever2013importance,ma2018quasi}, reinforcement learning \cite{vieillard2020momentum,korkmaz2020nesterov,haarnoja2018soft}, knowledge distillation \cite{tian2019contrastive,lee2022prototypical}, and recent semi-supervised learning \cite{tarvainen2017mean,cai2021exponential,li2021momentum}. Recently, EMA has also been applied in modern self-supervised learning frameworks. Amongst the seminal works, MoCo \cite{he2020momentum,chen2020improved,chen2021mocov3}, DINO \cite{caron2021emerging}, ReSSL \cite{zheng2021ressl}, and BYOL \cite{grill2020bootstrap} are examples that use the EMA in the target encoder to prevent model collapse \cite{grill2020bootstrap,caron2021emerging}, consistent negative samples \cite{he2020momentum,chen2020improved}, or boosting performance as in MoCo-v3 \cite{chen2021mocov3}. These EMA-based frameworks contain two branches (as shown in Fig. \ref{fig:architecture}): the first branch is an encoder that allows back-propagation \cite{NEURIPS2019_pytorch} during training, which refers to as \textit{online} encoder (student). The second branch is \textit{target} encoder (teacher), which is constructed by a momentum encoder which its parameters are dynamically updated via the \textit{online} encoder via Eq. \ref{eq:ema_update}. Besides the success of momentum-based SSL, our work is the first to investigate and resolve the representation gap between the teacher and student.


\textbf{Momentum-Free Self-Supervised Learning.} The momentum-free SSL approaches do not employ EMA in their frameworks but avoid collapse by using a contrastive loss (SimCLR \cite{chen2020simple}), a predictor and stop gradient (SimSiam \cite{chen2021exploring}), or the cross-correlation loss (Barlow Twins \cite{zbontar2021barlow}). 
Although these momentum-free approaches exhibit inferior performance in classification \ie in ImageNet-1K \cite{grill2020bootstrap,caron2021emerging,chen2021mocov3} or other downstream tasks \cite{chen2020improved,wang2020DenseCL,caron2021emerging} compared to the momentum-based approaches, they are still used in practice for certain cases thanks to their simple design, \ie removing the need for storing the parameters and one more forward of the momentum encoder \cite{madaan2022representational}.


\textbf{Representation Gap.} 
There have been early works trying to reduce the representation gap between the student and teacher networks in order to maximize the performance of the student model. In Knowledge Distillation (KD) \cite{hinton2015distilling}, the knowledge from a larger and better performing model (teacher) is used to generate the soft targets for a smaller student model, hence reducing the distribution gap between the two models. The other works of self-knowledge distillation (self-KD) try to use students themselves as teachers. In \cite{xu2019data}, the self-KD model is trained to reduce the distance between features extracted from two separate distorted versions of an image by a KL divergence loss. The self-training in \cite{tarvainen2017mean} may be close to our work where the distance between outputs of the teacher and student is minimized to improve generalizability. There are three points that the mean teacher model in \cite{tarvainen2017mean} is different from our work. First, the mean teacher approach works in a supervised learning paradigm where our proposed \textit{intra-momentum} works in a self-supervised manner, \ie without any labels involved. Second, \cite{tarvainen2017mean} minimizes the distribution distance or prediction of labels with MSE loss on softmax outputs while \textit{intra-momentum} is trained to minimize the representation gap between teacher and student with cosine similarity loss (\ie no softmax applied). And third, in \cite{tarvainen2017mean}, the teacher and student have different inputs with injected noises $\eta$ and $\eta'$. By contrast, our proposed \textit{intra-momentum} uses exactly the same input for both student and teacher models as Fig. \ref{fig:architecture}b).
\begin{figure}[!htbp]
  \centering
  \includegraphics[width=1.0\linewidth]{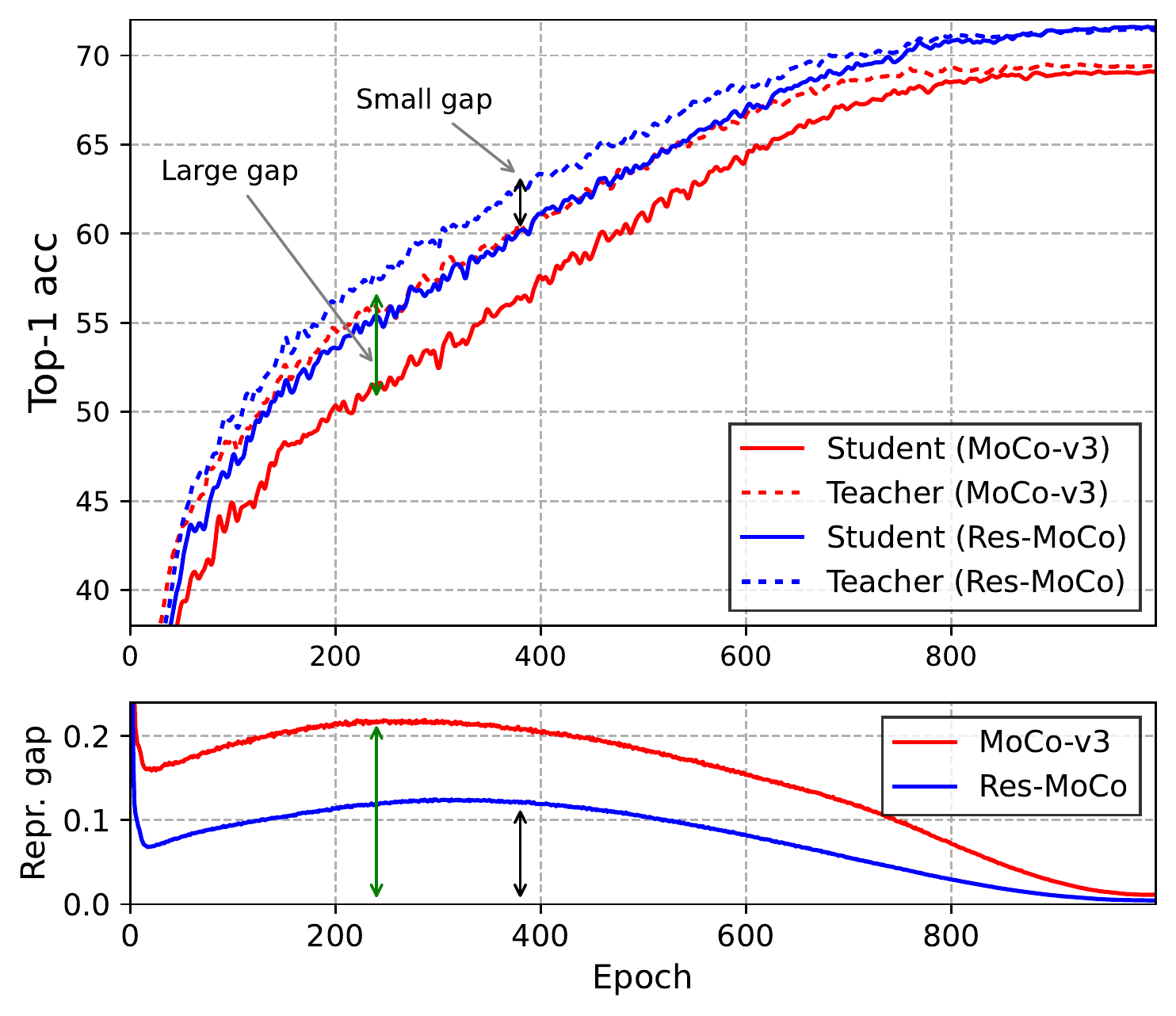}
  \vspace{-12pt}
  \caption{(Bottom) Intra representation gap between teacher and student on CIFAR-100 (measured by cosine similarity as Eq. \ref{eq:cosine_loss}) and (top) their corresponding performance gap. The proposed Res-MoCo significantly reduces the intra-representation gap and consequentially their performance gap is further narrowed. }
  \label{fig:compare_gap_teacher_student_cifar100}
  \vspace{-10pt}
\end{figure}

As discussed in ~\cite{caron2021emerging}, momentum-based SSL frameworks \cite{he2020momentum,caron2021emerging,grill2020bootstrap,chen2021mocov3} have the same form as KD.
For these SSLs, we notice that two different augmented images (positive pair) are fed separately into the \textit{teacher} and \textit{student} to learn the augmentation-invariant representation. During training, a loss function (either cross-entropy \cite{caron2021emerging}, contrastive loss \cite{he2020momentum,chen2021mocov3,zheng2021ressl}, or a simple cosine similarity \cite{grill2020bootstrap}) is applied in their outputs to minimize their gap. We refer to this gap as \textit{inter-representation gap} since they use different inputs and refer to the momentum in these frameworks as \textit{inter-momentum}.
By contrast, in this work, we focus on the representation gap between the teacher and the student for a given exactly the same input image. We refer to this gap as \textit{intra-representation gap} which can be minimized by the proposed \textit{intra-momentum}. 
Compared to the existing \textit{inter momentum}, the \textit{intra-momentum} has unique properties: \textit{First}, it is augmentation agnostic. \textit{Second}, different from the traditional inter-momentum that is integrated into the existing SSL loss, the proposed \textit{intra-momentum} is decoupled from the existing SSL loss so that it can be flexibly plugged into any other SSL frameworks (including both momentum-based and momentum-free ones) for boosting performance.
\section{Method}
\label{method}
We first present the background of SSL. After that, we analyze how \textit{inter-momentum} is applied in the existing SSL frameworks, \ie MoCo-v3 as our baseline. Then we introduce the preliminaries of the proposed \textit{intra-momentum} to narrow the \textit{intra-representation gap}. Finally, we introduce the objective function of the proposed method Res-MoCo, that consists of both \textit{inter-momentum} and \textit{intra-momentum}.

\subsection{Background}
Given an unlabeled image $\mathbf{x}\in \mathbb{R}^{H\times W\times 3}$, two random augmentations $\mathbf{x}_1\in \mathbb{R}^{H\times W\times 3}$ and $\mathbf{x}_2 \in \mathbb{R}^{H\times W\times 3}$ are generated and formed as a positive pair, the augmentations from the other images in the current mini-batch are treated as the negative samples \cite{chen2020simple,chen2020improved}. The two augmented images $\mathbf{x}_1$ and $\mathbf{x}_2$ are separately fed into two different encoders, \ie \textit{student} and \textit{teacher} to train the \textit{student} model to learn the invariant-augmentation representation\footnote{To avoid any ambiguity, we use the following concepts interchangeably: (\textit{online encoder} \cite{grill2020bootstrap} or \textit{student} \cite{caron2021emerging});
(\textit{target encoder} \cite{grill2020bootstrap} or \textit{momentum encoder} \cite{he2020momentum,chen2021mocov3} or \textit{teacher} \cite{caron2021emerging}).}.
We consider an online encoder (student), \ie $\mathbb{E}$ with a backbone $f$ (\eg ResNet-50), projector $g$ \cite{chen2020simple,he2020momentum} and may be followed by a predictor $q$ \cite{chen2021mocov3,grill2020bootstrap}. The target encoder (teacher), \ie $\mathbb{E}_m$ ($f_m,g_m,q_m$) has the same architecture as the student network. The subscript $m$ beside each character denotes momentum. The parameters of the student network are updated by the standard backpropagation \cite{NEURIPS2019_pytorch} while the teacher's parameters are momentum updated as follows \cite{grill2020bootstrap,chen2021mocov3}:
\begin{equation}
    \label{eq:ema_update}
    \xi \leftarrow \beta \xi + (1-\beta)\theta,
\end{equation}
where $\theta$ and $\xi$ are the parameters of the student $\mathbb{E}$ and teacher $\mathbb{E}_m$, respectively. A constant $\beta\in(0,1)$ is the momentum coefficient which is often chosen with a value of 0.99 for short training (200 epochs) \cite{grill2020bootstrap,chen2021exploring,he2020momentum}. In longer training, \ie 1000 epochs for full convergence, SSL methods often use the higher momentum value, \ie $\beta=0.996$ \cite{grill2020bootstrap,chen2021exploring,caron2021emerging,chen2021mocov3}. The presented framework is commonly used in most EMA-based SSL approaches. We specify the MoCo-v3 design as our baseline in the next section.

\subsection{Inter-Momentum to Minimize the Inter-Representation Gap in SSL}
MoCo \cite{he2020momentum} is the first work to introduce EMA for self-supervised contrastive learning and has become a groundbreaking and highly recognized framework. Its latest version, \ie MoCo-v3 \cite{chen2021mocov3} employs the best practices in the SSL area. Specifically, MoCo-v3 consists of the online encoder $\mathbb{E}$ with ($f,g,q$) and the target encoder $\mathbb{E}_m'$ with ($f_m,g_m$). Note that there is no predictor on the target encoder. Two crops $\mathbf{x}_1$ and $\mathbf{x}_2$ are embedded by $\mathbb{E}$ and $\mathbb{E}_m'$ to have the outputs $q$ (query) and $k_m$ (key). The objective function of MoCo-v3 \cite{chen2021mocov3} is adopted by InfoNCE loss \cite{oord2018representation}:
\begin{equation}
    \label{eq:infoNCE}
    \mathcal{L}_{\text{ctr}} = -\log \frac{\exp(q{\cdot}k^+_m/\tau)}{\exp(q{\cdot}k^+_m/\tau) + \sum\limits_{k^-_m}\exp(q{\cdot}k^-_m/\tau)},
\end{equation}
where ($\cdot$) denotes cosine similarity, $k_m^+$ is the output of $\mathbb{E}_m'$ for the augmentation of a same image as the query $q$ (positive sample), $k_m^-$ is the negative samples of $q$. Symbol $\tau$ is a temperature hyper-parameter \cite{chen2020simple}, $q$ and $k$ are $l_2$-normalized \cite{chen2021mocov3}. For every sample $\mathbf{x}$ in the current mini-batch, the above loss is symmetrized as follows MoCo-v3 \cite{chen2021mocov3}:
\begin{equation}
    \label{eq:inter_momentum}
    \mathcal{L}_{\text{Inter-M}} = \frac{1}{2}\left( \mathcal{L}_{\text{ctr}} (q_1,k_{2,m}) + \mathcal{L}_{\text{ctr}} (q_2,k_{1,m}) \right),
\end{equation}
here we put ``Inter-M'' to the subscript to denote that momentum used in MoCo-v3 (and all existing SSL frameworks) to construct the teacher and student that uses two different augmented images as inputs. Obviously, ``inter-momentum'' (Inter-M) in Eq. \ref{eq:inter_momentum} is designed to minimize the distance between two different inputs (inter-representation gap), \ie making $\mathbf{x}_1$ and $\mathbf{x}_2$ closer to learn the invariant-augmentation representations during the training process.

Next, we introduce a \textit{residual momentum} (``intra-momentum'') which is designed to have teacher and student use exactly the same augmented image input. It is proposed to minimize the distance between the outputs of the teacher and student for the same input (intra-representation gap).

\subsection{Intra-Momentum to Minimize the Intra-Representation Gap in SSL}
Previous works, \ie MoCo-v3 in the last section, only focus on Inter-M to minimize the gap between the teacher and student's outputs from two different augmented images while ignoring the potential gap between the teacher and student from the same augmented image. 

As shown in Fig. \ref{fig:compare_gap_teacher_student_cifar100}, such an intra-representation gap causes a big gap in their performance, which prevents the student model from learning good representation. To this end, we propose to measure the intra-representation gap between teacher and student models using the exact same input via \textit{intra-momentum} (Intra-M). 
Without changes in the architecture of MoCo-v3, we consider the online encoder $\mathbb{E}(f,g,q)$ and the momentum encoder $\mathbb{E}_m(f_m,g_m,q_m)$. MoCo-v3 uses the output of $g_m$ as the target (asymmetry), but Intra-M uses the output of $q_m$ (symmetry). We compute the intra-representation gap as follows:
\begin{equation}
    \label{eq:intra_gap}
    \mathcal{L}_{\text{Intra-gap}} = \mathcal{D}(q,q_m),
\end{equation}
where $\mathcal{D}$ is a distance function. In this paper, we consider three choices for the distance function. \textit{First}, we use the negative cosine similarity function (\textit{default}) as follows \cite{grill2020bootstrap}:
\begin{equation}
    \label{eq:cosine_loss}
    \mathcal{D}(q,q_m)_{\text{cosine}} = \Vert q - q_m \Vert_2^2 = 2 - 2 \cdot (q \cdot q_m),
\end{equation}
where $\Vert . \Vert$ denotes the $\ell_2$-norm. Vectors $q$ and $q_m$ are $\ell_2$-normalized. 
\textit{Second}, we ablate the other choice of distance function with the entropy function $H$ as used in knowledge distillation \cite{hinton2015distilling,caron2021emerging}:
\begin{equation}
    \label{eq:entropy_loss}
    \mathcal{D}(q,q_m)_{\text{CE}} = H(q,q_m) = - P(q)\log(P(q_m)),
\end{equation}
where $P(x)$ is the softmax output of a vector $x\in \mathbb{R}^{K}$: $P(x)^{(i)} = \frac{\exp(x)^{(i)}/\tau_{s}}{\sum_{k=1}^K\exp(x)^{(k)}/\tau_{s}}$, $\tau_{s}$ is a temperature parameter which the common choice in KD is \{3,4,5\} \cite{hinton2015distilling}. Here, $q$ and $q_m$ are not $\ell_2$-normalized.
And \textit{third}, we use the mean square error (MSE) to the softmax outputs of the student and teacher as in \cite{tarvainen2017mean}:
\begin{equation}
    \label{eq:mse_loss}
    \mathcal{D}(q,q_m)_{\text{MSE}} = \frac{1}{2}(q-q_m)^2,
\end{equation}
where $q$ and $q_m$ are not $\ell_2$-normalized but softmax-normalized. Note that $q$ and $q_m$ are the outputs of the online and momentum encoder, respectively, which come from the exact same image augmentation input. Our experiments show that Intra-M using cosine similarity performs the best compared to MSE or CE in top-1 linear accuracy. This suggests that cosine similarity for \textit{intra-momentum} is more suitable in self-supervised contrastive learning. We also tried the asymmetric design for Intra-M, \ie $\mathcal{L}_{\text{Intra-gap}} = \mathcal{D}(q,g_m)$ however, it performs worse than $\mathcal{L}_{\text{Intra-gap}} = \mathcal{D}(q,q_m)$. This suggests that the asymmetry does not benefit Intra-M. Finally, following $\mathcal{L}_{\text{Inter-M}}$ in Eq. \ref{eq:inter_momentum}, we use a symmetrized loss $\mathcal{L}_{\text{Intra-M}}$ for Intra-M as follows:
\begin{equation}
    \label{eq:intra_momentum}
    \mathcal{L}_{\text{Intra-M}} = \frac{1}{2} \left( \mathcal{L}_{\text{Intra-gap}}(q_1,q_{1,m}) + \mathcal{L}_{\text{Intra-gap}}(q_2,q_{2,m}) \right).
\end{equation}

\subsection{Res-MoCo: Objective Function}
\begin{figure}[!tbp]
  \centering
  \subfloat[Top-1 linear accuracy (\%)] {\includegraphics[width=0.5\linewidth]{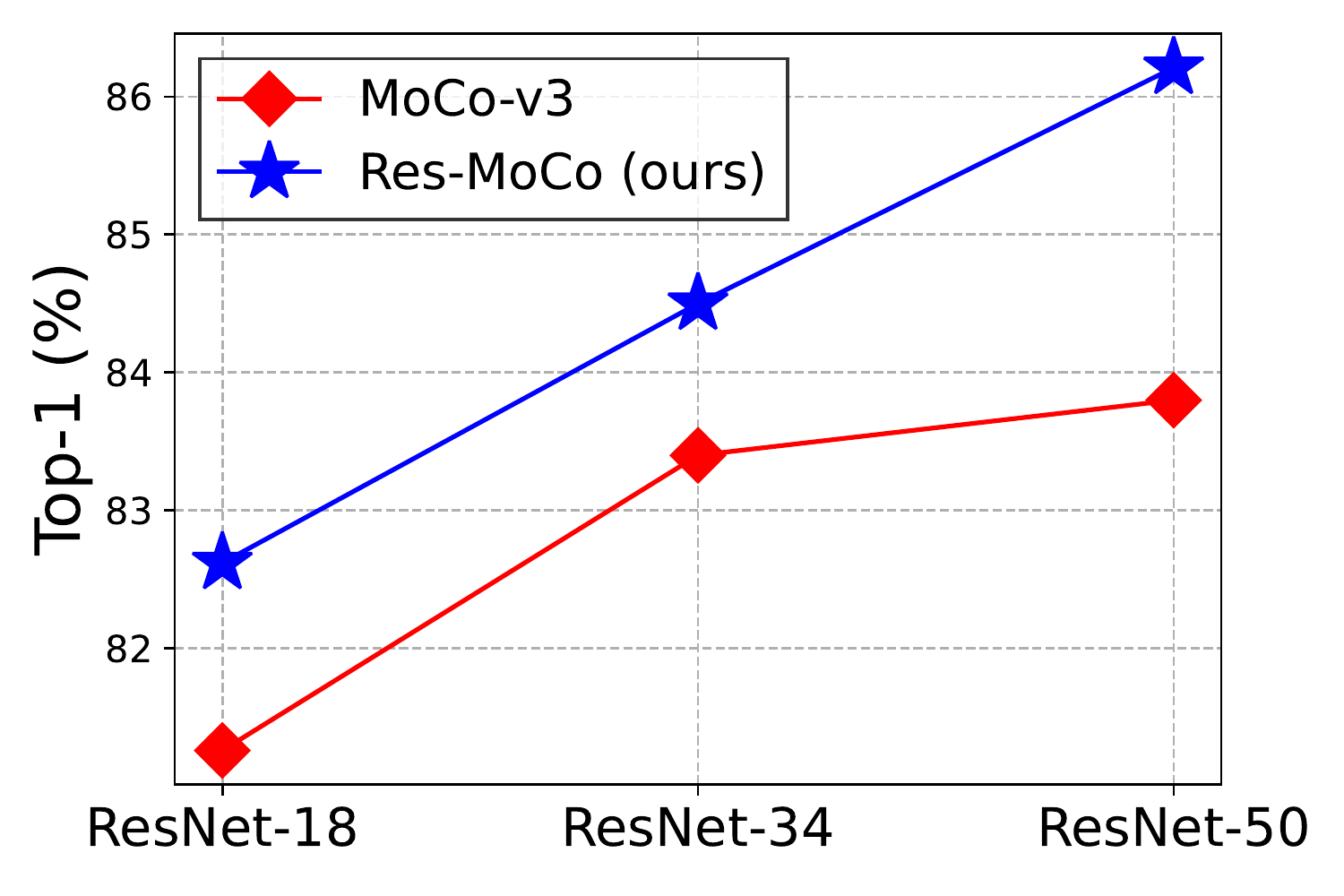}}
  \hfill
  \subfloat[Top-5 linear accuracy (\%)] {\includegraphics[width=0.5\linewidth]{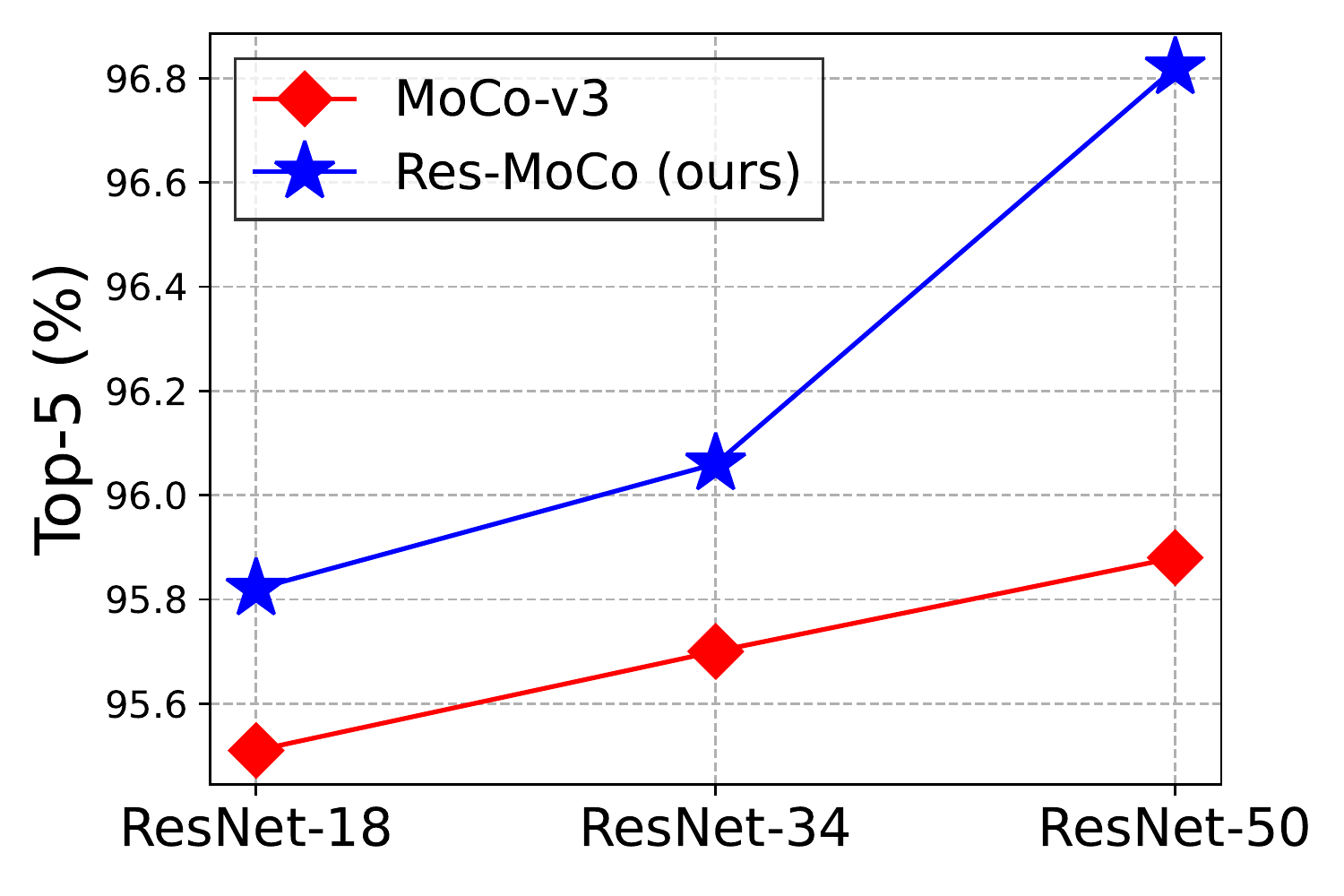}}
  \caption{Impact of intra-momentum (blue line). We compare \textit{MoCo-v3} and the proposed \textit{Res-MoCo} on ImageNet-100, pretraining 1000 epochs. Our method \textit{Res-MoCo} (with intra-momentum) enjoys significant improvements over the baseline \textit{MoCo-v3} for various network architectures. We report the linear classification accuracy top-1 and top-5 on the test set. It is best viewed in color. }
  \label{fig:acc_arch_acc1_acc5}
\end{figure}

We adopt MoCo-v3 as our baseline. Our final method is named \textbf{Res-MoCo} (residual momentum contrastive learning), whose loss is 
a combination of the two momentum losses (Eq. \ref{eq:inter_momentum} and Eq. \ref{eq:intra_momentum}) for joint optimization as follows:
\begin{equation}
    \label{eq:inter_intra}
    \mathcal{L}_{\text{Res-MoCo}} = \mathcal{L}_{\text{Inter-M}} + \mathcal{L}_{\text{Intra-M}}.
\end{equation}
By default, Res-MoCo uses cosine similarity (Eq. \ref{eq:cosine_loss}) for $\mathcal{L}_{\text{Intra-M}}$. Eq. \ref{eq:inter_intra} shows that Res-MoCo is designed to narrow the overall representation gap between the student and teacher, including \textit{inter-representation gap} (existing MoCo-v3) and \textit{intra-representation gap} (this work). We conduct an analysis on the impact of each component as in the ablation section. Compared to MoCo-v3 \cite{chen2021mocov3}, our Res-MoCo achieves superior performance for all architectures and datasets as highlighted in Fig. \ref{fig:acc_arch_acc1_acc5} and Fig. \ref{fig:ema_whole_part}.


\section{Experiments}
\subsection{Setup and Evaluation}
\noindent
\textbf{Datasets.}
We use CIFAR-10 (10 classes) \cite{krizhevsky2010cifar}, CIFAR-100 (100 classes) \cite{krizhevsky2010cifar}, ImagetNet-100 (100 classes) \cite{tian2020contrastive}, and ImageNet-1K (1000 classes) \cite{krizhevsky2012imagenet} for classification and VOC07+12 \cite{Everingham10} and COCO2017 \cite{lin2014microsoft} for object detection. We consider different network architectures to train MoCo-v3 and Res-MoCo such as ResNet-18 \cite{he2016deep}, ResNet-34 \cite{he2016deep}, ResNet-50 \cite{he2016deep}, and ViT-S \cite{dosovitskiy2021an}. 

\noindent
\textbf{Pre-training Setup.}
The open-source SSL library solo-learn \cite{turrisi2021sololearn} and its heavily tuned hyper-parameters are used for the experiments. We keep the same settings/parameters for MoCo-v3 and Res-MoCo. The encoder is pre-trained on the training set of each dataset without labels. Except for other mention, we train MoCo-v3 and Res-MoCo for 1000 epochs on CIFAR-10, CIFAR-100, and ImageNet-100 with ResNet-18 backbone and 200 epochs for large ImageNet-1K with ResNet-50 backbone.\footnote{As a common practice, we use the starting momentum coefficient $\beta=0.99$ for 200 epochs and $\beta=0.996$ for 1000 epochs. The value $\beta$ is increasing to 1.0 using a cosine scheduler as in \cite{chen2021mocov3,grill2020bootstrap}. More setup details are provided in the supplementary.}

\noindent
\textbf{Evaluation.}
Following prior works \cite{chen2021mocov3,grill2020bootstrap,chen2020simple}, we evaluate the proposed SSL framework in linear classification and transfer learning to object detection. We evaluate the pre-trained representations by training a linear classifier on frozen representations with the corresponding test set \cite{turrisi2021sololearn,grill2020bootstrap,chen2021mocov3}. For object detection, we use the ResNet-50 that is pre-trained by MoCo-v3 and Res-MoCo on ImageNet-1K to initialize the Faster R-CNN detector and finetune with a standard 2x schedule as in \cite{chen2020improved,he2020momentum,chen2021exploring} for each dataset, VOC07+12 object detection, and COCO object detection.

\begin{table}[t]
    \begin{center}
    \resizebox{1.0\hsize}{!}{
    \begin{tabular}{cccccc}
    \toprule
    {\bf Method} & Inter-M & Intra-M & Top-1 (\%) & Top-5 (\%) & KNN-1 (\%) \\
    \midrule
    W-MSE \cite{ermolov2021whitening} & - & - & 88.67 & 99.68 & - \\
    SwAV \cite{caron2020unsupervised} & - & - & 89.17 & 99.68 & - \\
    SimCLR \cite{chen2020simple} & - & - & 90.74 & 99.75 & - \\
    SimSiam \cite{chen2021exploring} & - & - & 90.51 & 99.72 & - \\
    VICReg \cite{bardes2022vicreg} & - & - & 92.07 & 99.74 & - \\
    NNCLR \cite{dwibedi2021little} & - & - & 91.88 & 99.78 & - \\
    Barlow Twins \cite{zbontar2021barlow} & - & - & 92.10 & 99.73 & - \\
    ReSSL \cite{zheng2021ressl} & \checkmark & - & 90.63 & 99.62 & - \\
    DINO \cite{caron2021emerging} & \checkmark & - & 89.52 & 99.71 & - \\
    BYOL \cite{grill2020bootstrap} & \checkmark & - & 92.58 & 99.79 & 88.84 \\
    MoCo-v3 \cite{chen2021mocov3} & \checkmark & - & 93.10 & 99.80 & 89.12 \\
    \midrule
    \rowcolor{gg}
    \bf Res-MoCo (ours)  & - & \checkmark & \bf 93.53 $_{+0.43}$ & \bf 99.88 $_{+0.08}$ & \bf 90.78 $_{+1.66}$ \\
    \rowcolor{gg}
    \bf Res-MoCo (ours)  & \checkmark & \checkmark & \bf 93.71 $_{+0.61}$ & \bf 99.84 $_{+0.04}$ & \bf 90.78 $_{+1.66}$ \\
    \bottomrule
    \end{tabular}}
    \end{center}
    \caption{\textbf{CIFAR-10}. Comparison of \textit{MoCo-v3} and \textit{Res-MoCo}. All methods are trained in full 1000 epochs using the same settings, such as 1 GPU, batch 256, and ResNet-18. We employ the results for other SSL methods from the official solo-learn library \cite{turrisi2021sololearn}.}
    \label{tab:mocov3_intra_cifar10}
\end{table}

\subsection{Results}
\noindent
\textbf{Linear Classification.}
In Tab. \ref{tab:mocov3_intra_cifar10} and Tab. \ref{tab:mocov3_intra_cifar100}, we report the results on CIFAR-10 and CIFAR-100 (size of 32x32) compared Res-MoCo to its baseline MoCo-v3 and the other state-of-the-art methods. On CIFAR-10, Res-MoCo outperforms MoCo-v3 for all metrics with 93.71\% (+0.61\%), 99.84\% (+0.04\%), and 90.78\% (+1.66\%) on top-1, top-5, and KNN-1 accuracy, respectively, and surpassing all considered state-of-the-art methods. On CIFAR-100, the improvement is more significant when Res-MoCo achieves state-of-the-art performance and outperforms baseline MoCo-v3 with 2.82\%, 2.25\%, and 3.52\% on top-1, top-5, and KNN-1, respectively.
Tab. \ref{tab:compare_mocov3_intra_inter_imagenet100}, Tab. \ref{tab:compare_mocov3_intra_inter_imagenet1k_Res50} and Tab. \ref{tab:compare_imagenet100_vitS_res50} report the results for the large-scale dataset (size of 224x224), ImageNet-100, and ImageNet-1K, respectively. Overall, Res-MoCo (Res-BYOL) with the presence of \textit{intra-momentum} shows a clear improvement with the highest accuracy among all methods. These experiments demonstrate the effectiveness of \textit{intra-momentum} for SSL frameworks.

\noindent
\textbf{Transfer Learning to Object Detection.}
In Tab. \ref{tab:compare_object_detection_Res50}, the quality of representations is evaluated by transferring them to the other tasks, including VOC \cite{Everingham10} object detection and COCO \cite{lin2014microsoft} object detection. The pre-trained models are finetuned end-to-end in the target datasets using the public code \cite{turrisi2021sololearn,he2020momentum}. MoCo-v3 and Res-MoCo are trained for 200 epochs in ImageNet-1K, showing competitive results among the leading methods when transferring the representations beyond the ImageNet task. Note that the other methods are trained with more epochs, \ie 800 or 1000 epochs. 
In Tab. \ref{tab:compare_object_detection_Res50}, Res-MoCo shows a clear improvement for all metrics, and all datasets compared to the MoCo-v3 baseline and surpasses all other competitors in COCO for AP, AP50, and AP75. This suggests better quality representations learned by Res-MoCo for tasks beyond classification.
\begin{table}[t]
    \begin{center}
    \resizebox{1.0\hsize}{!}{
    \begin{tabular}{cccccc}
    \toprule
    {\bf Method} & Inter-M & Intra-M & Top-1 (\%) & Top-5 (\%) & KNN-1 (\%) \\
    \midrule
    W-MSE \cite{ermolov2021whitening} & - & - & 61.33 & 87.26 & - \\
    SwAV \cite{caron2020unsupervised} & - & - & 64.88 & 88.78 & - \\
    SimCLR \cite{chen2020simple} & - & - & 65.78 & 89.04 & 58.52 \\
    SimSiam \cite{chen2021exploring} & - & - & 66.04 & 89.97 & 59.01 \\
    VICReg \cite{bardes2022vicreg} & - & - & 68.54 & 90.83 & - \\
    NNCLR \cite{dwibedi2021little} & - & - & 69.62 & 91.52 & - \\
    Barlow Twins \cite{zbontar2021barlow} & - & - & 70.84 & 92.04 & 62.35 \\
    ReSSL \cite{zheng2021ressl} & \checkmark & - & 65.92 & 89.91 & 59.05 \\
    DINO \cite{caron2021emerging} & \checkmark & - & 66.76 & 88.63 & 56.58 \\
    BYOL \cite{grill2020bootstrap} & \checkmark & - & 70.06 & 92.12 & 61.83 \\
    MoCo-v3 \cite{chen2021mocov3} & \checkmark & - & 68.83 & 90.07 & 60.75 \\
    \midrule
    \rowcolor{gg}
    \bf Res-BYOL (ours)  & - & \checkmark & \bf 71.27 $_{+1.21}$ & \bf 92.47 $_{+0.35}$ & \bf 63.17 $_{+1.34}$ \\
    \rowcolor{gg}
    \bf Res-BYOL (ours)  & \checkmark & \checkmark & \bf 71.43 $_{+1.37}$ & \bf 92.54 $_{+0.42}$ & \bf 62.34 $_{+0.54}$ \\
    \midrule
    \rowcolor{gg}
    \bf Res-MoCo (ours)  & - & \checkmark & \bf 70.56 $_{+1.73}$ & \bf 92.47 $_{+2.40}$ & \bf 63.40 $_{+2.65}$ \\
    \rowcolor{gg}
    \bf Res-MoCo (ours)  & \checkmark & \checkmark & \bf 71.65 $_{+2.82}$ & \bf 92.32 $_{+2.25}$ & \bf 64.27 $_{+3.52}$ \\
    \bottomrule
    \end{tabular}}
    \end{center}
    \vspace{-14pt}
    \caption{\textbf{CIFAR-100}. Comparison of \textit{MoCo-v3} and \textit{Res-MoCo}. All methods are trained in 1000 epochs using the same settings, such as 1 GPU, batch 256, and ResNet-18. We employ the results for the other SSL methods from the official solo-learn library \cite{turrisi2021sololearn}.}
    \label{tab:mocov3_intra_cifar100}
\end{table}
\begin{table}[!htbp]
    \begin{center}
    \resizebox{1.0\hsize}{!}{
    \begin{tabular}{cccccc}
    \toprule
    {\bf Method} & Inter-M & Intra-M & Top-1 (\%) & Top-5 (\%) & KNN-1 (\%) \\
    \midrule
    SimCLR \cite{chen2020simple} & - & - & 78.46 & - & 73.22 \\
    MoCo-v2 \cite{chen2020improved} & \checkmark & - & 79.98 & - & 74.56 \\ 
    BYOL \cite{grill2020bootstrap} & \checkmark & - & 81.46 & 95.26 & 75.34 \\
    MoCo-v3 \cite{chen2021mocov3} & \checkmark & - & 81.26 & 95.51 & 75.28 \\
    MoCo-v3 w/o Mom. & - & -  & 78.71 & 94.32 & 73.24 \\
    \midrule
    \rowcolor{gg}
    \bf Res-BYOL (ours) & \checkmark & \checkmark & \bf 82.58$_{+1.12}$ & \bf 95.44$_{+0.18}$ & \bf 75.76$_{+0.42}$ \\
    \midrule
    \rowcolor{gg}
    \bf Res-MoCo (ours) & - & \checkmark  & \bf 81.66$_{+0.40}$ & \bf 95.86$_{+0.35}$ & \bf 76.24$_{+0.96}$ \\
    \rowcolor{gg}
    \bf Res-MoCo (ours) & \checkmark & \checkmark  & \bf 82.62$_{+1.36}$ & \bf 95.82$_{+0.31}$ & \bf 75.84$_{+0.56}$ \\
    \bottomrule
    \end{tabular}}
    \end{center}
    \vspace{-14pt}
    \caption{\textbf{ImageNet-100}. Compare \textit{MoCo-v3} and \textit{Res-MoCo}. Methods are trained for 1000 epochs with the same settings, 2 GPUs, batch 256, and ResNet-18. Mom. denotes ``momentum''.}
    \label{tab:compare_mocov3_intra_inter_imagenet100}
    \vspace{-10pt}
\end{table}
\begin{table}[!htbp]
    \begin{center}
    \resizebox{1.0\hsize}{!}{
    \begin{tabular}{ccccccc}
    \toprule
    \multicolumn{1}{c}{\bf Method} & Epoch & Inter-M & Intra-M & Architecture & Top 1 & Top-5  \\
    \midrule
    SimCLR \cite{chen2020simple} & 200 & - & - & ResNet-50 & 68.3 & - \\
    SwAV \cite{caron2020unsupervised} & 200 & - & - & ResNet-50 & 69.1 & - \\
    MoCo-v2 \cite{chen2020improved} & 200 & \checkmark & - & ResNet-50 & 69.9 & - \\
    SimSiam \cite{chen2021exploring} & 200 & - & - & ResNet-50 & 70.0 & - \\
    BYOL \cite{grill2020bootstrap} & 200 & \checkmark & - & ResNet-50 & 70.6 & - \\
    MoCo-v3 \cite{chen2021mocov3} & 200 & \checkmark & - & ResNet-50 & 71.0 & 90.0 \\
    \midrule
    \rowcolor{gg}
    \bf Res-MoCo (ours) & 200 & \checkmark & \checkmark & ResNet-50 & \bf 71.7 $_{+0.7}$ & \bf 90.4 $_{+0.4}$ \\
    \bottomrule
    \end{tabular}}
    \end{center}
    \vspace{-14pt}
    \caption{\textbf{ImageNet-1K}. Linear classification. Comparisons of MoCo-v3 \cite{chen2021mocov3} and Res-MoCo with the other approaches.}
    \label{tab:compare_mocov3_intra_inter_imagenet1k_Res50}
\end{table}

\begin{table}[t]
    \begin{center}
    \resizebox{1.0\hsize}{!}{
    \begin{tabular}{cccccccccc}
    \toprule
    \bf Method & Inter-M & Intra-M & Arch. & Top-1 & $\Delta_{top-1}$ & Top-5 & $\Delta_{top-5}$ & KNN-1 & $\Delta_{knn}$ \\
    \midrule
    MoCo-v3 \cite{chen2021mocov3} & \checkmark & - & ViT-S & 79.66 & - & 95.40 & - & 76.64 & - \\
    \rowcolor{gg}
    \bf Res-MoCo (ours) & \checkmark & \checkmark & ViT-S & \bf 79.94 & + 0.28 & \bf 95.78 & + 0.38 & \bf 77.46 & + 0.82  \\
    \midrule
    MoCo-v3 \cite{chen2021mocov3} & \checkmark & - & ResNet-18 & 81.26 & - & 95.51 & - & 75.28 & - \\
    \rowcolor{gg}
    \bf Res-MoCo (ours) & \checkmark & \checkmark & ResNet-18 & \bf 82.62 & + 1.36 & \bf 95.82 & + 0.31 & \bf 75.84 & + 0.56 \\
    \midrule
    MoCo-v3 \cite{chen2021mocov3} & \checkmark & - & ResNet-34 & 83.40 & - & 95.70 & - & 78.82 & - \\
    \rowcolor{gg}
    \bf Res-MoCo (ours) & \checkmark & \checkmark & ResNet-34 & \bf 84.50 & + 1.10 & \bf 96.06 & + 0.36 & \bf 79.74 & + 0.92 \\
    \midrule
    MoCo-v3 \cite{chen2021mocov3} & \checkmark & - & ResNet-50 & 83.80 & - & 95.88 & - & 72.80 & - \\
    \rowcolor{gg}
    \bf Res-MoCo (ours) & \checkmark & \checkmark & ResNet-50 & \bf 86.21 & + 2.41 & \bf 96.82 & + 0.94 & \bf 74.46 & + 1.66 \\
    \bottomrule
    \end{tabular}}
    \end{center}
    \vspace{-14pt}
    \caption{\textbf{ Different backbone architectures.} Comparison of MoCo-v3 and Res-MoCo on ImageNet-100. All methods are trained for 1000 epochs using 2 GPUs, batch size 256.}
    \label{tab:compare_imagenet100_vitS_res50}
    \vspace{-10pt}
\end{table}
\begin{table}[t]
    \begin{center}
    \resizebox{1.0\hsize}{!}{
    \begin{tabular}{cccccccc}
    \toprule
    \multirow{2}{*}{\bf Method} & \multirow{2}{*}{Epoch} &  & \multicolumn{1}{c}{VOC07+12} &  &  & COCO &  \\
     & & AP & AP50 & AP75 & AP & AP50 & AP75 \\
    \midrule
    Random Init. & - & 33.8 & 60.2 & 33.1 & 29.9 & 47.9 & 32.0 \\
    \midrule
    Pre-Train Sup. & 200 & 54.2 & 81.6 & 59.8 & 38.2 & 58.2 & 41.2 \\
    \midrule
    MoCo \cite{he2020momentum} & 200 & 55.9 & 81.5 & 62.6 & 38.5 & 58.3 & 41.6 \\
    SimCLR \cite{chen2020simple} & 1000 & 56.3 & 81.9 & 62.5 & 37.9 & 57.7 & 40.9 \\
    MoCo-v2 \cite{chen2020improved} & 800 & 57.4 & 82.5 & 64.0 & 39.3 & 58.9 & 42.5 \\
    Barlow Twins \cite{zbontar2021barlow} & 1000 & 56.8 & 82.6 & 63.4 & 39.2 & 59.0 & 42.5 \\
    VICReg \cite{bardes2022vicreg} & 1000 & - & 82.4 & - & 39.4 & - & - \\
    BYOL \cite{grill2020bootstrap} & 200 & 55.3 & 81.4 & 61.1 & 37.9 & 57.8 & 40.9 \\
    SwAV \cite{caron2020unsupervised} & 1000 & 56.1 & 82.6 & 62.7 & 38.4 & 58.6 & 41.3 \\
    \midrule
    MoCo-v3 \cite{chen2021mocov3} & 200 & 56.2 & 82.4 & 62.9 & 39.1 & 58.8 & 42.2 \\
    \rowcolor{gg}
    \bf Res-MoCo (ours) & 200 & 56.5$_{+0.3}$ & \bf 82.6$_{+0.2}$ & 63.0$_{+0.1}$ & \bf 39.5$_{+0.4}$ & \bf 59.2$_{+0.4}$ & \bf 42.9$_{+0.7}$ \\
    \bottomrule
    \end{tabular}}
    \end{center}
    \vspace{-14pt}
    \caption{\textbf{Object Detection} on VOC07+12 and COCO2017. Comparison of Res-MoCo and the existing methods using ResNet-50 pre-trained on ImageNet-1K for transfer learning. Results MoCo-v3 and Res-MoCo are run three times and taken average.}
    \label{tab:compare_object_detection_Res50}
\end{table}

\section{Ablation Study}
\subsection{Gap Reduction Functions}
The simplest choice to reduce the representation gap between the teacher and student is \textit{cosine similarity} loss. We also ablate the other choices of the distance function such as \textit{cross-entropy} (CE) (Eq. \ref{eq:entropy_loss}) or mean square error (MSE) (Eq. \ref{eq:mse_loss}). We find that in SSL, the cosine similarity loss performs the best compared to the cross entropy in KD \cite{hinton2015distilling} or MSE in self-training \cite{tarvainen2017mean}. This indicates that cosine similarity with \textit{intra-momentum} is more suitable for learning representations in contrastive learning frameworks (Tab. \ref{tab:distance_loss}).
\begin{table}[!htbp]
    \begin{center}
    \resizebox{0.75\hsize}{!}{
    \begin{tabular}{c|ccc}
    Method & Top 1 & Top-5 & KNN-1 \\
    \hline
    MoCo-v3 \cite{chen2021mocov3} & 68.83 & 90.07 & 60.75 \\
    \hline
    Res-MoCo w/ MSE (Eq. \ref{eq:mse_loss}) \cite{tarvainen2017mean} & 68.73 & 90.92 & 60.61 \\
    Res-MoCo w/ CE (Eq. \ref{eq:entropy_loss}) \cite{hinton2015distilling} & 69.72 & 91.85 & 63.14 \\
    Res-MoCo w/ Cosine & \bf 71.65 & \bf 92.32 & \bf 64.27 \\
    \end{tabular}}
    \end{center}
    \vspace{-16pt}
    \caption{Choices for $\mathcal{L}_{\text{Intra-M}}$. Performance on CIFAR-100, trained for 1000 epochs when all models are fully converged.}
    \label{tab:distance_loss}
\end{table}

\subsection{Intra-Representation Gap Matters}
\begin{figure}[!htbp]
  \centering
  \subfloat[CIFAR-10 (ResNet-18)] {\includegraphics[width=0.5\linewidth]{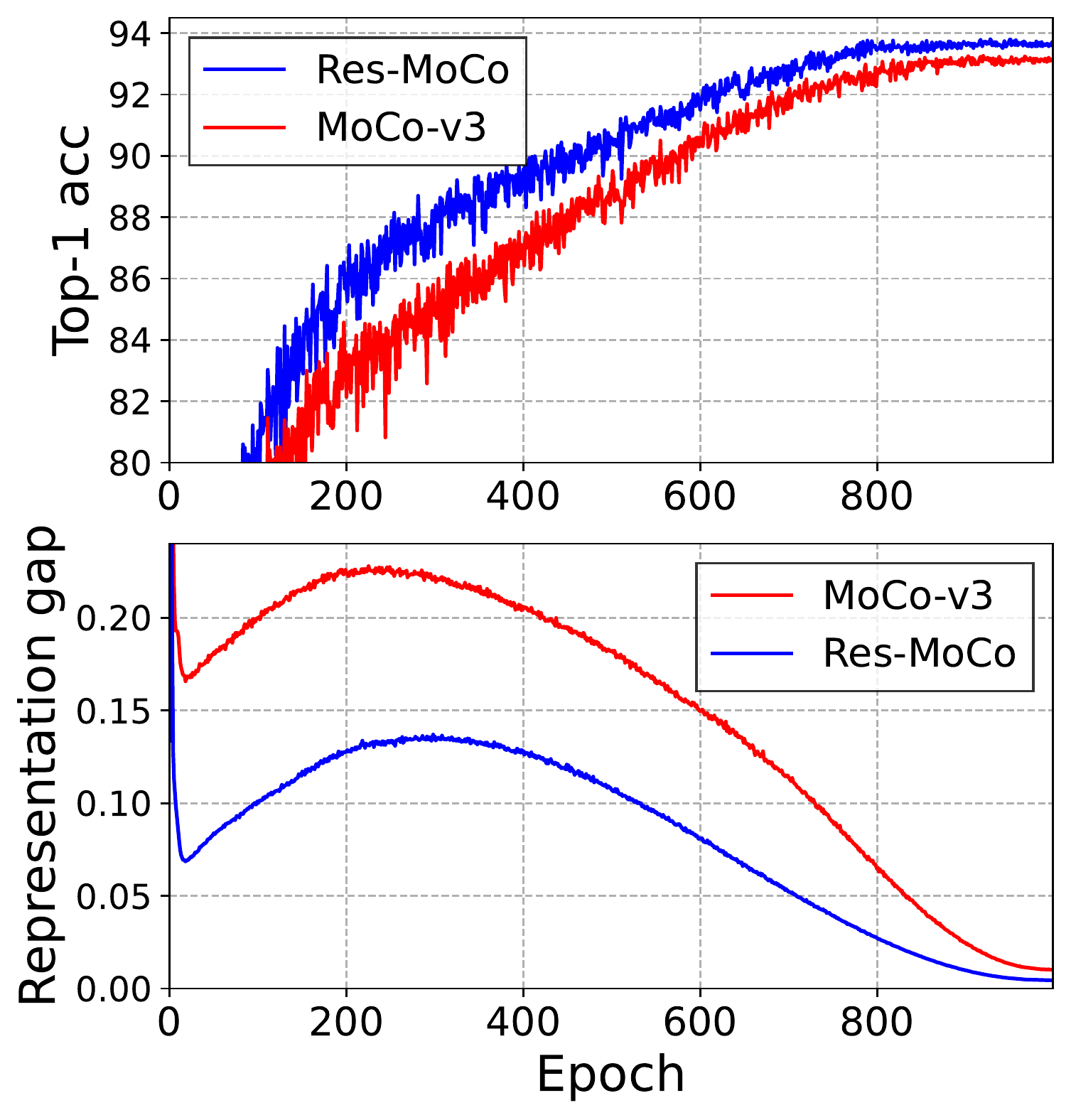}}
  \hfill
  \subfloat[CIFAR-100 (ResNet-18)] {\includegraphics[width=0.5\linewidth]{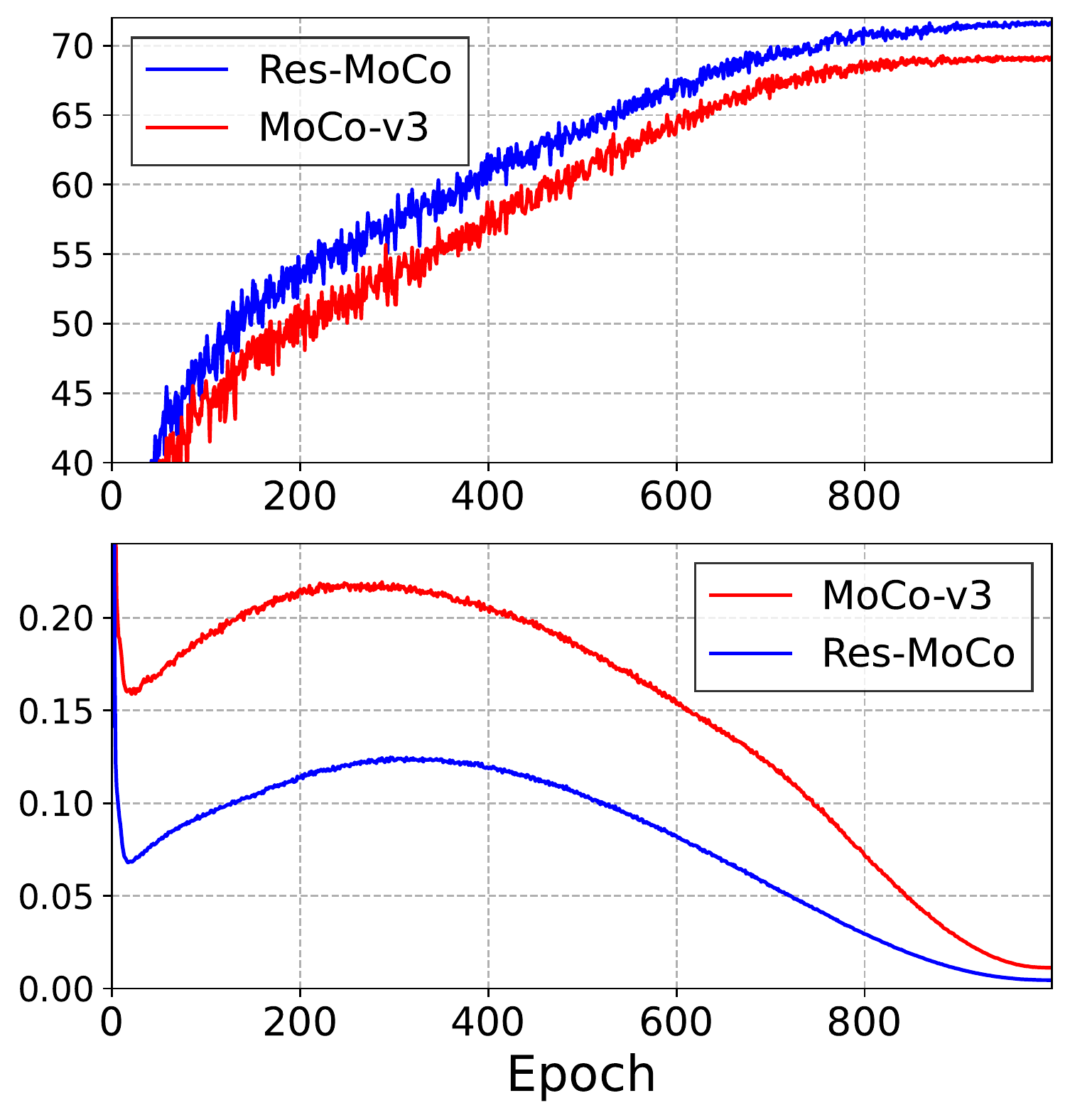}}
  \caption{Compare the representation gap between teacher and student for MoCo-v3 and Res-MoCo on CIFAR-10, CIFAR-100. We report the corresponding top-1 test accuracy of the student model.}
  \label{fig:compare_acc1_intra_cifar10}
  \vspace{-7pt}
\end{figure}

\begin{figure}[!htbp]
  \vspace{-7pt}
  \centering
  \subfloat[ImageNet-100 (ResNet-18)] {\includegraphics[width=0.5\linewidth]{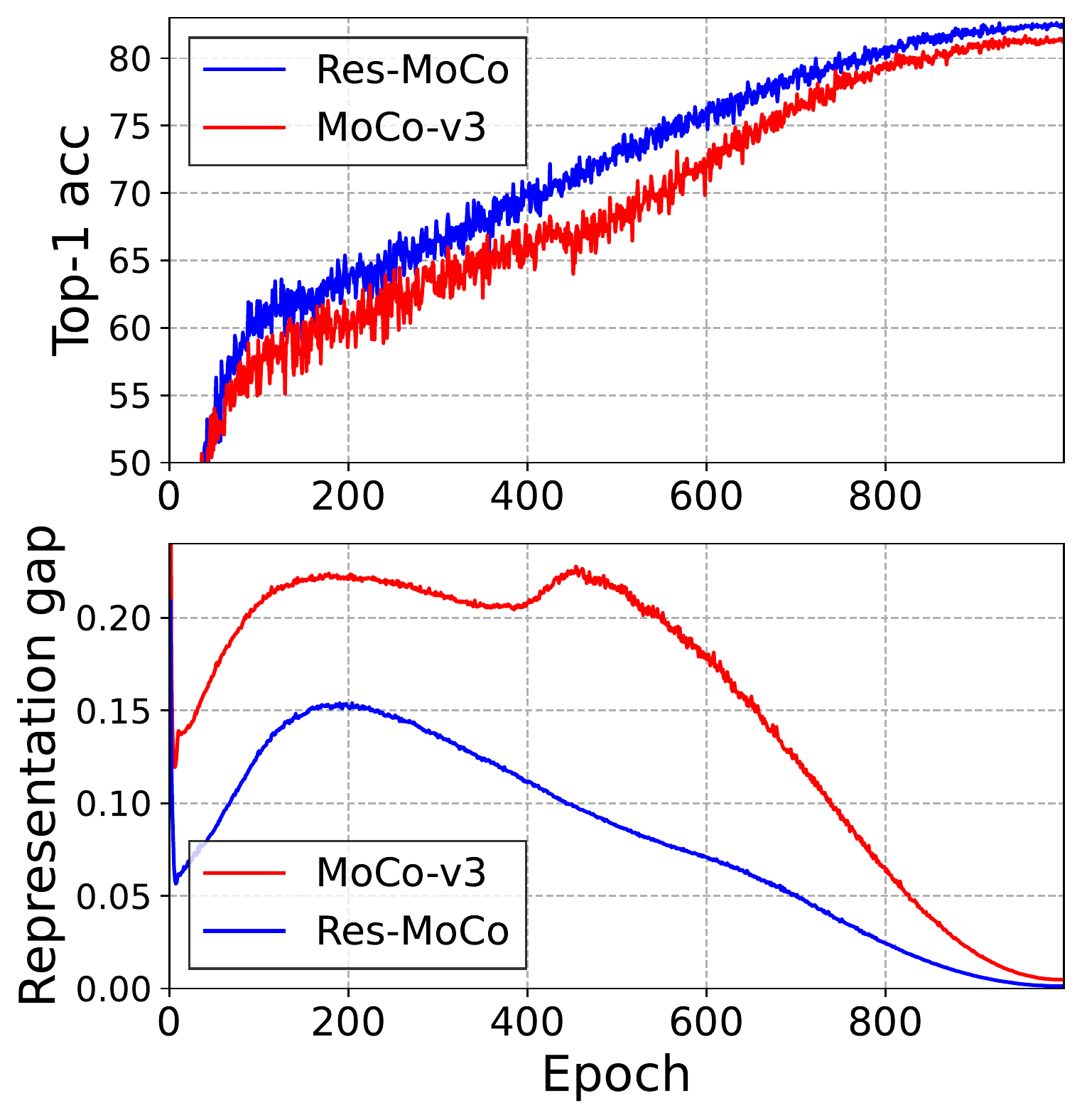}}
  \hfill
  \subfloat[ImageNet-1K (ResNet-50)] {\includegraphics[width=0.5\linewidth]{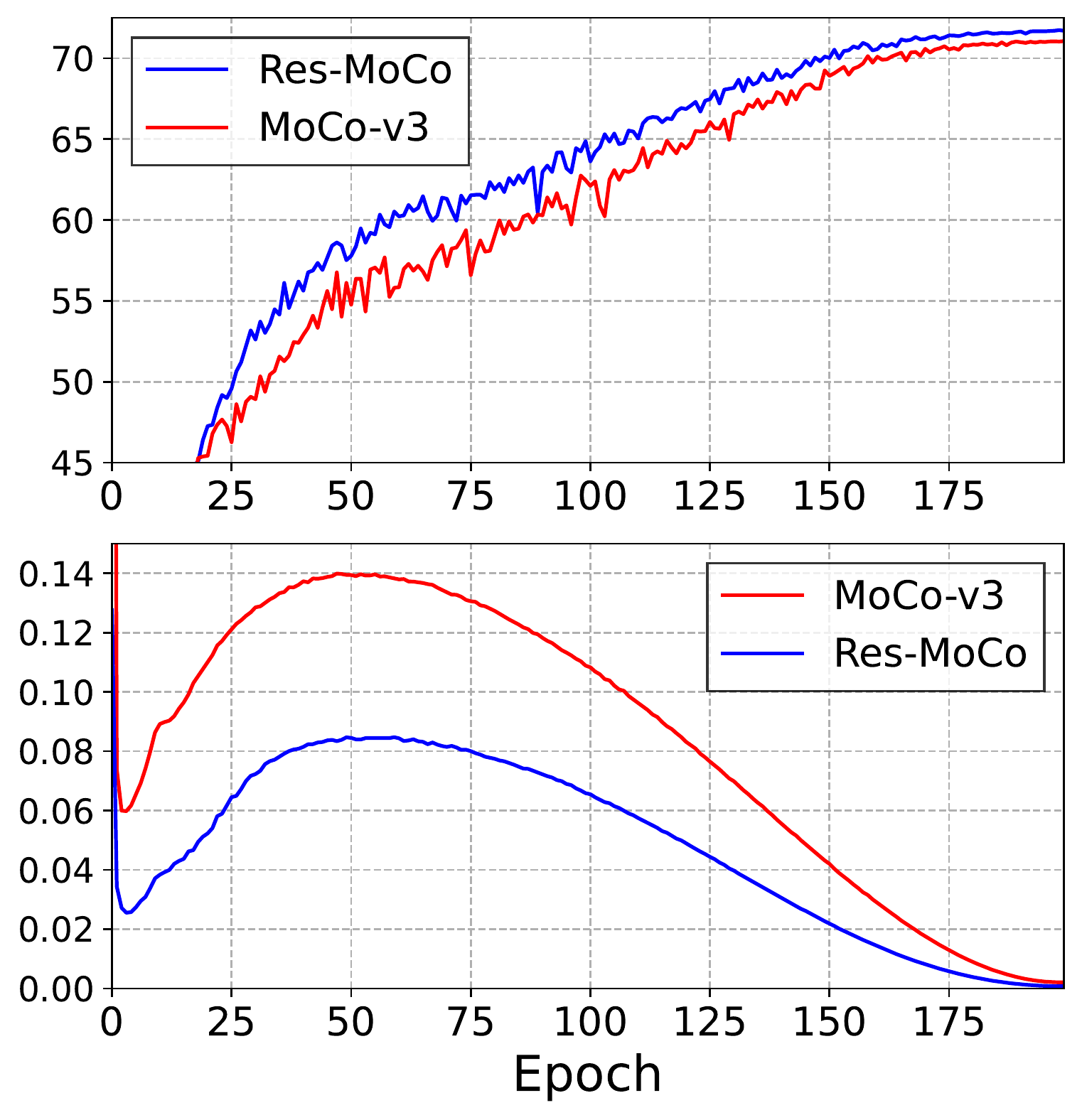}}
  \caption{Compare representation gap between teacher and student for MoCo-v3 and Res-MoCo on ImageNet-100, ImageNet-1K. We report the corresponding top-1 test accuracy of the student model.}
  \label{fig:compare_acc1_intra_IN100}
\end{figure}

\begin{figure}[t]
  \centering
  \includegraphics[width=0.9\linewidth]{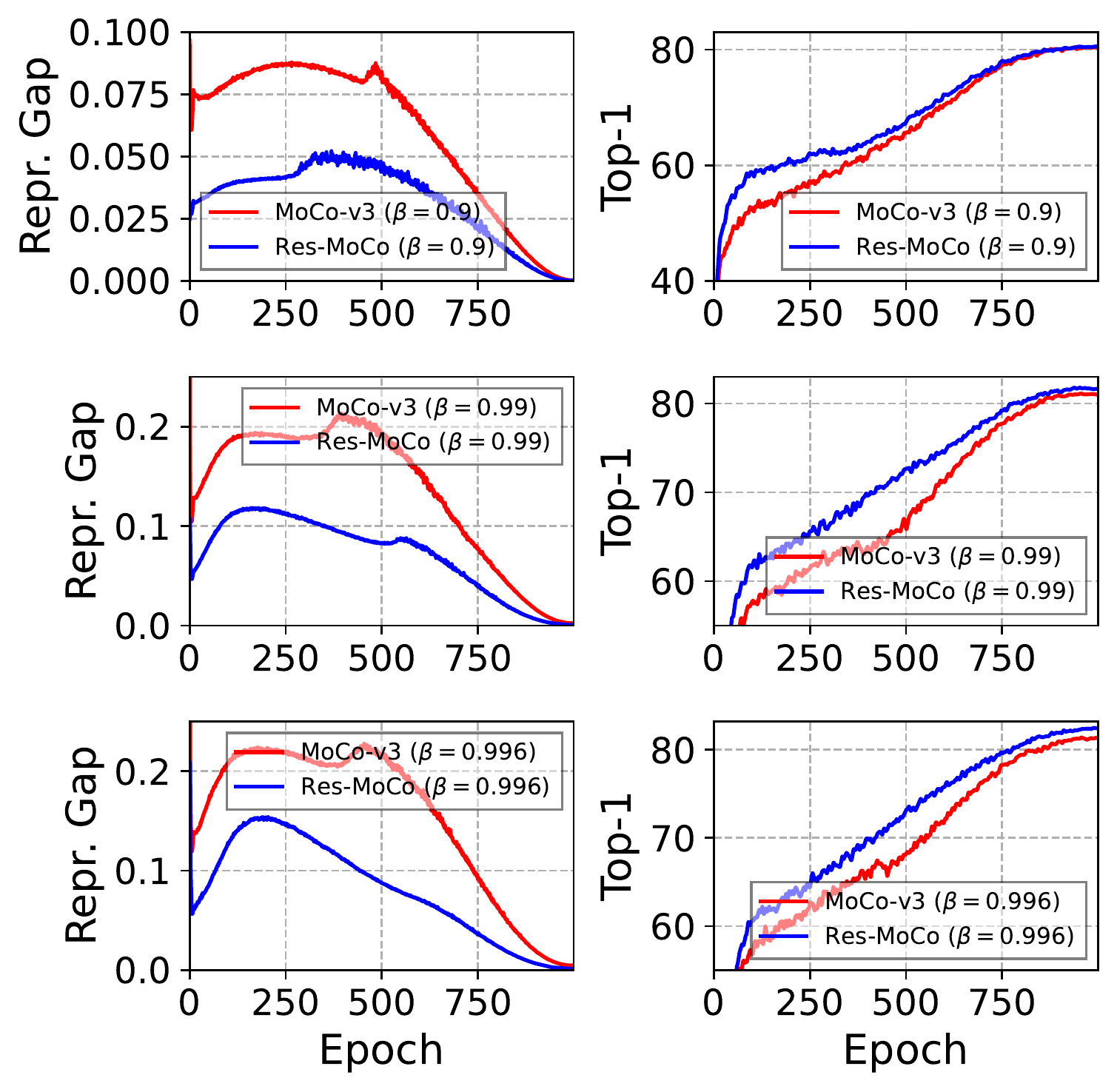}
  \vspace{-12pt}
  \caption{Representation gap between teacher and student encoders of MoCo-v3 and Res-MoCo with different momentum update speeds $\beta$ on ImageNet-100. It is measured by the negative cosine similarity of the current training batch as in Eq. \ref{eq:cosine_loss}.}
  \label{fig:different_beta_mocov3_resmoco}
  \vspace{-16pt}
\end{figure}

\textbf{Different Datasets.}
We provide a learning curve and quantitative metric to clearly show the intra-representation gap when training Res-MoCo with and without \textit{residual momentum} (MoCo-v3). As shown in Fig. \ref{fig:compare_acc1_intra_cifar10} and Fig. \ref{fig:compare_acc1_intra_IN100}, for both considered datasets, Res-MoCo manages to significantly reduce the representation gap between the teacher and student, which correspondingly boosting performance.
\textbf{Behaviour Gap with Different Momentum Speeds.}
In this part, we ablate the potential gap between the student and teacher encoder in different momentum update speeds $\beta=\{0.9,0.99,0.996 \}$. As shown in Fig. \ref{fig:different_beta_mocov3_resmoco}, for 1000 epochs, $\beta=0.996$ gives the best performance for MoCo-v3; however, it also shows the largest representation gap. Although the network architectures of the teacher and student are identical, the EMA updates the weight from the student to the teacher with a very slow momentum speed, \ie $\beta=0.996$ \cite{grill2020bootstrap,caron2021emerging,chen2021exploring,pham2022pros}, therefore their weights are still quite different. In Eq. \ref{eq:ema_update}, with $\beta=0.996$, only 0.4\% of the student's weight is counted to compute the new weight for the teacher. Fig. \ref{fig:different_beta_mocov3_resmoco} shows that there is a considerable gap in their produced representations of MoCo-v3, which we identify potentially prevents the SSL model learn good representations. We find that the intra-representation gap keeps increasing and only narrows at the end of training. At this point, with a nearly zero learning rate (cosine decay), the model converged, and the overall weight was insignificantly updated. 
Reducing the representation gap with \textit{intra-momentum} makes a clear effect of forcing the student model to learn to perform as closely as the teacher possible during training, hence improving both models together. Note that the teacher model is dynamically updated with EMA from the student; therefore, the better the student model learned, the better teacher models are updated, and vice versa.

\begin{figure*}[t] 
  \centering
  \includegraphics[width=1.0\linewidth]{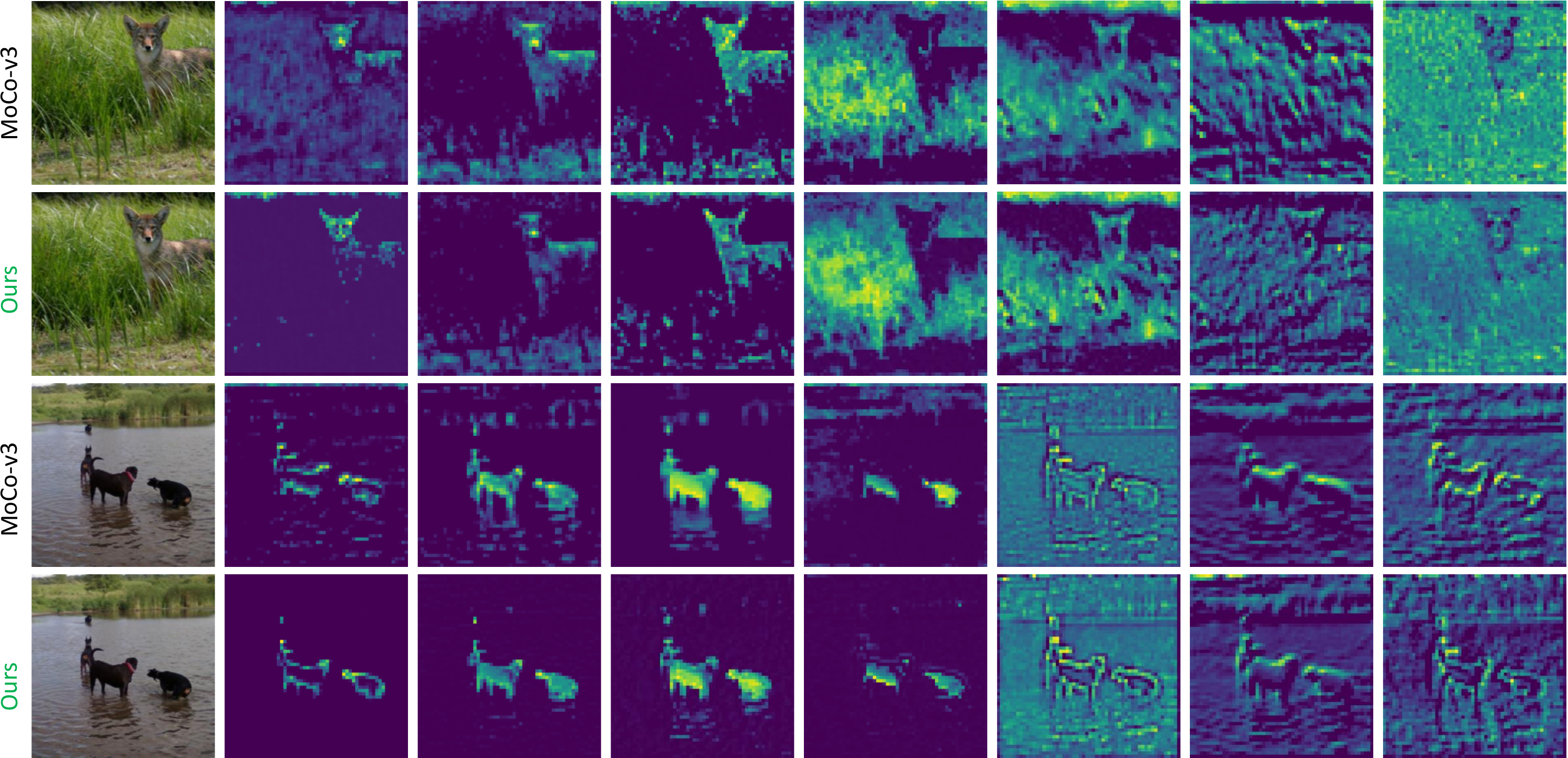}
  \caption{Features learned by MoCo-v3 \cite{chen2021mocov3} and Res-MoCo (ours) with \textit{intra-momentum}. Two image samples are from the test set of ImageNet-100. We visualize the seven most visually meaningful feature maps of the last convolutional layer in the first block of ResNet-50. It is best viewed in color and zoom-in. More feature maps and quantitative results can be found in the supplementary.}
  \label{fig:feature_maps}
\end{figure*}
\begin{figure*}[t] 
  \centering
  \includegraphics[width=1.0\linewidth]{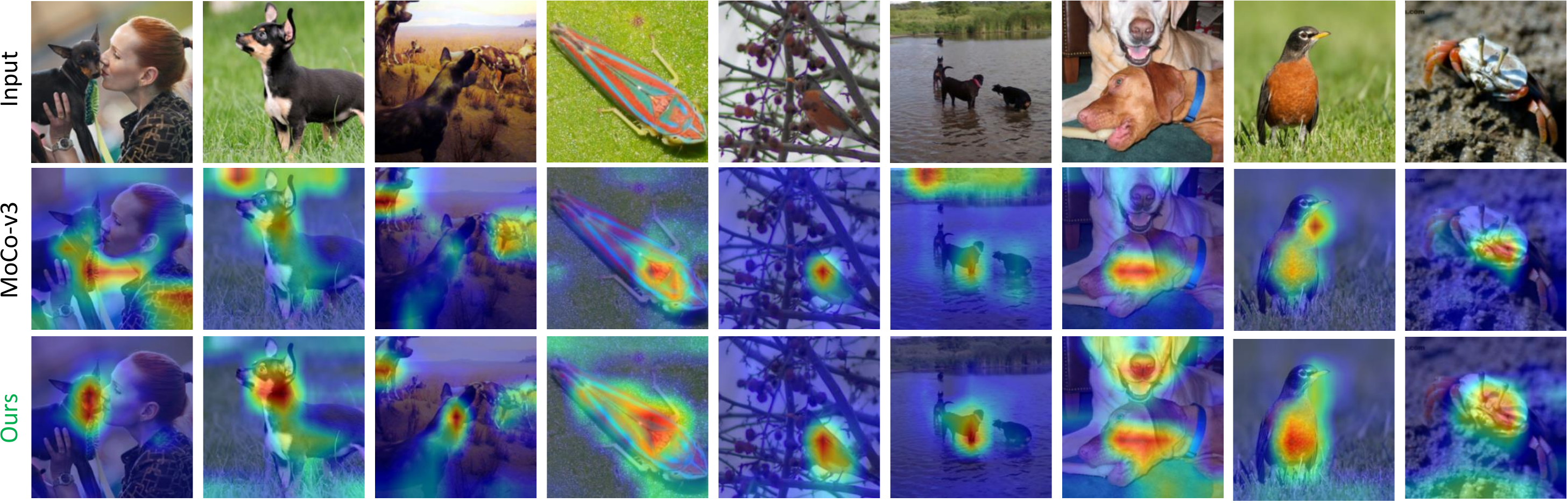}
  \caption{Comparison of GradCAM learned by MoCo-v3 and Res-MoCo (ours). The image samples are from the test set of ImageNet-100. The first row is the original image, the second row is the heat map by MoCo-v3, and the third row is the heat map produced by Res-MoCo (ours). It clearly shows that the heat map produced by Res-MoCo is more accurate than MoCo-v3. It is best viewed in color.}
  \label{fig:grad_cam}
\end{figure*}

\subsection{Impact of Inter-M and Intra-M}
\noindent
\textbf{MoCo-v3 w/ and w/o Inter-Momentum.}
When removing the momentum encoder in MoCo-v3, the loss function for self-supervised contrastive learning becomes as follows:
\vspace{-6pt}
\begin{equation}
    \label{eq:CL}
    \mathcal{L}_{\text{CL}}  = -\log \frac{\exp(q{\cdot}k^+/\tau)}{\exp(q{\cdot}k^+/\tau) + \sum_{k^-}\exp(q{\cdot}k^-/\tau)}.
\end{equation}
We emphasize that the key $k_m$ (with subscript $m$) in Eq. \ref{eq:inter_momentum} comes from the EMA encoder, but in Eq. \ref{eq:CL}, the key $k$ comes from the same online encoder with \textit{stop gradient} \cite{chen2021exploring} (\ie teacher $\leftarrow$ student). As shown in Tab. \ref{tab:mocov3_intra_cifar100},\ref{tab:compare_mocov3_intra_inter_imagenet100} and Fig. \ref{fig:huge_representation_gap}, MoCo-v3 w/o EMA yields 78.71\% top-1, MoCo-v3 gives 81.26\%. The Inter-M helps to boost 2.55\% improvement. Next, we inspect the impact of Intra-M and its combination.

\noindent
\textbf{Contrastive Learning with Intra-Momentum.}
We ablate the case when removing the existing Inter-M but replacing it with Intra-M. Specifically, new loss contains two components as \textit{intra-momentum} and \textit{CL} terms:
$\mathcal{L}_{\text{CL-intra\_M}} = \mathcal{L}_{\text{CL}} + \mathcal{L}_{\text{intra\_M}}.$
Tab. \ref{tab:compare_mocov3_intra_inter_imagenet100} shows that Intra-M boosts CL to 81.66\% (+2.95\%), which is alone even better than the existing Inter-M. Combining inter-M and intra-M furthers boosts to 82.65\%. These results suggest that Intra-M can be used as an alternative way to the existing Inter-M to get the same or better benefit of momentum. The proposed Intra-M is also an easy way to plug into any other SSL framework (EMA-based or EMA-free frameworks) since Intra-M is decoupled from the existing SSL loss, while Inter-M is not.
\subsection{Visualizations}
To understand how Res-MoCo consistently outperforms MoCo-v3 for most tasks, we analyze their pre-trained models with different aspects. First, we compare the feature maps learned by MoCo-v3 and the proposed Res-MoCo. Second, we consider the GradCAM heat map, a powerful tool in deep learning for model interpretation \cite{selvaraju2017grad}. The ResNet-50 backbone trained on ImageNet-100 for 1000 epochs in the above experiments is chosen for comparisons.

\textbf{Learned feature maps.} Fig. \ref{fig:feature_maps} shows the quantitative results of our method Res-MoCo against MoCo-v3. We can observe in 4 first columns that feature maps produced by Res-MoCo are cleaner when it removes background noises compared to those produced by MoCo-v3. Removing the unnecessary parts in the picture and focusing only on the objects is the key to achieving good performance in object recognition and object detection.

\textbf{GradCAM attention map.} As shown in Fig. \ref{fig:grad_cam}, the proposed Res-MoCo produces the attention heatmap that focuses on the object more accurately than MoCo-v3 for various examples. For example, in the first image, when the model is called backpropagation for the dog class, the attention heat map of MoCo-v3 points out the dog and other irrelevant parts. By contrast, the Res-MoCo heat map can accurately focus on the dog's face only. These visualizations visually demonstrate the significant improvements of Res-MoCo over baseline MoCo-v3 on the object detection task with VOC and COCO benchmark datasets.

\section{Conclusions}
\label{conclusions}
In this paper, we investigate that there is still a big representation gap between the teacher and student in the existing momentum-based SSL frameworks during training. This overlooked gap potentially prevents the model from learning good representations. To this end, we propose Res-MoCo with \textit{intra-momentum} to reduce such a gap. As a result, the student model is trained to match closely with the teacher's output and yield better performance. Extensive experiments demonstrate that our method improves state-of-the-art approaches' performance with visible margins. With our findings, we hope that the representation gap between teacher and student should be observed more carefully in future SSL works for better representation learning.


{\small
\bibliographystyle{ieee_fullname}
\bibliography{egbib}
}


\appendix
\noindent
{\large \textbf{Supplementary Materials} }
\section{Implementation Details}
We implement the proposed Res-MoCo on top of the baseline MoCo-v3 \cite{chen2021mocov3} as the designed code in the open-source SSL library \cite{turrisi2021sololearn} for backbone, projector, predictor, etc. The details about the settings of MoCo-v3 frameworks as described in the tables below. We just used the same hyper-parameters for both MoCo-v3 and Res-MoCo. We use NVIDIA RTX A6000 in our experiments. For ImageNet-100, training 1000 epochs with 2 GPUs, batch size 256, it took 1.5 days (ResNet-18) to complete one experiment. ImageNet-1K took 2.5 days to train 200 epochs using 4 GPUs (ResNet-50) with batch size 1024 for one experiment. CIFAR-10 and CIFAR-100 took 1.75 days to train 1000 epochs using 1 GPU (ResNet-18) with batch size 256 for each experiment. For one experiment, it took 3 days to evaluate the trained SSL models for object detection with the COCO dataset using 8 GPUs of Titan Xp NVIDIA. Finally, Pascal VOC07+12 took 9h for evaluation using 8 GPUs of Titan Xp NVIDIA.

\subsection{Augmentations}
\textbf{CIFAR-10 and CIFAR-100.} We use the following augmentations Tab. \ref{tab:augmentation_cifar}, which are the same as the heavily tuned setup as the default of the solo-learn library \cite{turrisi2021sololearn}.
\begin{table}[!htbp]
    \begin{center}
    \resizebox{1.0\hsize}{!}{
    \begin{tabular}{ccc}
    \toprule
    \bf Parameter & $T_1$ & $T_2$ \\
    \midrule
    \midrule
    min scale & 0.2 & 0.2 \\
    crop size & 32x32 & 32x32 \\
    random crop probability & 1.0 & 1.0 \\
    brightness adjustment max intensity & 0.4 & 0.4 \\
    contrast adjustment max intensity & 0.4 & 0.4 \\
    saturation adjustment max intensity & 0.2 & 0.2 \\
    hue adjustment max intensity & 0.1 & 0.1 \\
    gaussian blurring probability & 1.0 & 0.1 \\
    solarization probability & 0.0 & 0.2 \\
    \bottomrule
    \end{tabular}}
    \end{center}
    \vspace{-14pt}
    \caption{Parameters used for image augmentations in CIFAR-10 and CIFAR-100. }
    \label{tab:augmentation_cifar}
    \vspace{-14pt}
\end{table}

\textbf{ImageNet-100 and ImageNet-1K.} The settings for the augmentations in SSL training are in Tab. \ref{tab:augmentation_imagenet}.
\begin{table}[!htbp]
    \begin{center}
    \resizebox{1.0\hsize}{!}{
    \begin{tabular}{ccc}
    \toprule
    \bf Parameter & $T_1$ & $T_2$ \\
    \midrule
    \midrule
    crop size & 224x224 & 224x224 \\
    random crop probability & 1.0 & 1.0 \\
    brightness adjustment max intensity & 0.4 & 0.4 \\
    contrast adjustment max intensity & 0.4 & 0.4 \\
    saturation adjustment max intensity & 0.2 & 0.2 \\
    hue adjustment max intensity & 0.1 & 0.1 \\
    gaussian blurring probability & 1.0 & 0.1 \\
    solarization probability & 0.0 & 0.2 \\
    \bottomrule
    \end{tabular}}
    \end{center}
    \vspace{-14pt}
    \caption{Parameters used for image augmentations in ImageNet-100 and ImageNet-1K. }
    \label{tab:augmentation_imagenet}
    \vspace{-14pt}
\end{table}

\subsection{Hyper-Parameters}
In the proposed Res-MoCo with \textit{intra-momentum}, there are no additional hyper-parameters compared to MoCo-v3. In all settings, we just use the same hyperparameters set. For contrastive loss of MoCo-v3 \cite{chen2021mocov3}, temperature $\tau=0.2$ is used as default \cite{chen2021mocov3,turrisi2021sololearn}.
The settings for CIFAR-10 and CIFAR-100 are shown in Tab. \ref{tab:hyperparameters_cifar}, for ImageNet-100 and ImageNet-1K in Tab. \ref{tab:hyperparameters_imagenet100} and Tab. \ref{tab:hyperparameters_imagenet1k}, respectively.
\begin{table}[!htbp]
    \begin{center}
    \resizebox{1.0\hsize}{!}{
    \begin{tabular}{cc}
    \toprule
    \bf Parameter & Value \\
    \midrule
    \midrule
    max epoch & 1000 \\
    the number of GPU & 1 \\
    optimizer & lars \\
    learning rate $\eta$ for LARS & 0.02 \\
    exclude bias norm lars & yes \\
    pre-training learning rate scheduler & warmup cosine \\
    learning rate & 0.3 \\
    weight decay & 1e-6 \\
    batch size & 256 \\
    projection output dimension & 256 \\
    projection hidden dimension & 4096 \\
    prediction hidden dimension & 4096 \\
    temperature $\tau$ & 0.2 \\
    base tau momentum $\beta$ & 0.996 \\
    final tau momentum $\beta$ & 1.0 \\
    scheduler for momentum $\beta$ & cosine increases 0.996 to 1.0 \\
    \bottomrule
    \end{tabular}}
    \end{center}
    \vspace{-14pt}
    \caption{Parameters used for networks trained in CIFAR-10 and CIFAR-100 datasets. }
    \label{tab:hyperparameters_cifar}
\end{table}

\begin{table}[!htbp]
    \begin{center}
    \resizebox{1.0\hsize}{!}{
    \begin{tabular}{cc}
    \toprule
    \bf Parameter & Value \\
    \midrule
    \midrule
    max epoch & 1000 \\
    the number of GPU & 2 \\
    optimizer & lars \\
    batch norm type & synchronization batch norm \\
    learning rate $\eta$ for LARS & 0.02 \\
    exclude bias norm lars & yes \\
    pre-training learning rate scheduler & warmup cosine \\
    learning rate & 0.3 \\
    weight decay & 1e-6 \\
    batch size & 256 (each GPU has 128) \\
    projection output dimension & 256 \\
    projection hidden dimension & 4096 \\
    prediction hidden dimension & 4096 \\
    temperature $\tau$ & 0.2 \\
    base tau momentum $\beta$ & 0.996 \\
    final tau momentum $\beta$ & 1.0 \\
    scheduler for momentum $\beta$ & cosine increases 0.996 to 1.0 \\
    \bottomrule
    \end{tabular}}
    \end{center}
    \vspace{-14pt}
    \caption{Parameters used for networks trained ImageNet-100. }
    \label{tab:hyperparameters_imagenet100}
\end{table}

\begin{table}[!htbp]
    \begin{center}
    \resizebox{1.0\hsize}{!}{
    \begin{tabular}{cc}
    \toprule
    \bf Parameter & Value \\
    \midrule
    \midrule
    max epoch & 200 \\
    the number of GPU & 4 \\
    optimizer & lars \\
    batch norm type & synchronization batch norm \\
    learning rate $\eta$ for LARS & 0.002 \\
    exclude bias norm lars & yes \\
    pre-training learning rate scheduler & warmup cosine \\
    learning rate & 0.45 \\
    weight decay & 1.5e-6 \\
    batch size & 1024 (each GPU has 256) \\
    projection output dimension & 256 \\
    projection hidden dimension & 4096 \\
    prediction hidden dimension & 4096 \\
    temperature $\tau$ & 0.2 \\
    base tau momentum $\beta$ & 0.99 \\
    final tau momentum $\beta$ & 1.0 \\
    scheduler for momentum $\beta$ & cosine increases 0.99 to 1.0 \\
    \bottomrule
    \end{tabular}}
    \end{center}
    \vspace{-14pt}
    \caption{Parameters used for networks trained ImageNet-1K. }
    \label{tab:hyperparameters_imagenet1k}
\end{table}

\section{More Ablation Study}

\begin{table*}[t]
    \begin{center}
    \resizebox{0.8\hsize}{!}{
    \begin{tabular}{cccccccccc}
    \toprule
    \bf Method & Inter-M & Intra-M & Arch. & Top-1 & $\Delta_{top-1}$ & Top-5 & $\Delta_{top-5}$ & KNN-1 & $\Delta_{knn}$ \\
    \midrule
    \midrule
    MoCo-v3 \cite{chen2021mocov3} $\tau=0.2$ & \checkmark & - & ResNet-50 & 83.80 & - & 95.88 & - & 72.80 & - \\
    \rowcolor{gg}
    \bf Res-MoCo (ours) $\tau=0.2$  & \checkmark & \checkmark & ResNet-50 & \bf 86.21 & + 2.41 & \bf 96.82 & + 0.94 & \bf 74.46 & + 1.66 \\
    \midrule
    MoCo-v3 \cite{chen2021mocov3} $\tau=1.0$ & \checkmark & - & ResNet-50 & 82.54 & - & 95.50 & - & 66.28 & - \\
    \bf Res-MoCo $\tau=1.0$ & \checkmark & \checkmark & ResNet-50 & \bf 84.64 & + 2.10 & \bf 96.28 & + 0.78 & \bf 71.12 & + 4.84 \\
    \bottomrule
    \end{tabular}}
    \end{center}
    \vspace{-10pt}
    \caption{\textbf{Different temperatures.} Comparison of MoCo-v3 and Res-MoCo on ImageNet-100. All methods are trained for 1000 epochs using the same settings run on 2 GPUs, batch size 256. It shows that Res-MoCo outperforms MoCo-v3 for both temperature settings.}
    \label{tab:compare_temperature}
\end{table*}

\begin{table*}[t]
    \begin{center}
    \resizebox{0.8\hsize}{!}{
    \begin{tabular}{cccccccccc}
    \toprule
    \bf Method & Epoch & Inter-M & Intra-M & Sim (\%) & $\Delta_{sim}$ & Top-1 & $\Delta_{top-1}$ & KNN-1 & $\Delta_{knn}$ \\
    \midrule
    \midrule
    MoCo-v3 \cite{chen2021mocov3} & 100 & \checkmark & - & 90.74 & - & 59.76 & - & 55.88 & - \\
    \bf Res-MoCo (ours) & 100 & \checkmark & \checkmark & \bf 94.04 & + 3.30 & \bf 64.21 & + 4.45 & \bf 61.89 & + 6.01 \\
    \midrule
    MoCo-v3 \cite{chen2021mocov3} & 200 & \checkmark & - & 90.15 & - & 60.87 & - & 56.70 & - \\
    \bf Res-MoCo (ours) & 200 & \checkmark & \checkmark & \bf 93.01 & + 2.86 & \bf 63.84 & + 2.97 & \bf 61.96 & + 5.26 \\
    \midrule
    MoCo-v3 \cite{chen2021mocov3} & 400 & \checkmark & - & 88.48 & - & 64.16 & - & 58.60 & - \\
    \bf Res-MoCo (ours) & 400 & \checkmark & \checkmark & \bf 93.90 & + 5.42 & \bf 69.88 & + 5.72 & \bf 66.21 & + 7.61 \\
    \midrule
    MoCo-v3 \cite{chen2021mocov3} & 800 & \checkmark & - & 96.88 & - & 80.15 & - & 68.57 & - \\
    \bf Res-MoCo (ours) & 800 & \checkmark & \checkmark & \bf 98.41 & + 1.53 & \bf 82.74 & + 2.59 & \bf 74.52 & + 5.95 \\
    \midrule
    MoCo-v3 \cite{chen2021mocov3} & 1000 & \checkmark & - & 99.84 & - & 83.80 & - & 72.80 & - \\
    \bf Res-MoCo (ours) & 1000 & \checkmark & \checkmark & \bf 99.95 & + 0.11 & \bf 86.21 & + 2.41 & \bf 74.46 & + 1.66 \\
    \bottomrule
    \end{tabular}}
    \end{center}
    \vspace{-10pt}
    \caption{Similarity of the online and momentum version of each image itself on \textbf{ImageNet-100}, \textbf{ResNet-50}. During training, Res-MoCo shows a much higher similarity for the image and its momentum version, which strongly corresponds to the higher performance in both linear top-1 (\%) and KNN accuracy (\%). We monitor similarity (Sim) by taking the average cosine similarity of each training batch. }
    \label{tab:CosineSim_imagenet100_res50}
\end{table*}

\begin{table*}[!htbp]
    \begin{center}
    \resizebox{0.8\hsize}{!}{
    \begin{tabular}{cccccccccc}
    \toprule
    \bf Method & Epoch & Inter-M & Intra-M & Sim (\%) & $\Delta_{sim}$ & Top-1 & $\Delta_{top-1}$ & KNN-1 & $\Delta_{knn}$ \\
    \midrule
    \midrule
    MoCo-v3 \cite{chen2021mocov3} & 100 & \checkmark & - & 89.65 & - & 54.62 & - & 54.98 & - \\
    \rowcolor{gg}
    \bf Res-MoCo (ours) & 100 & \checkmark & \checkmark & \bf 93.63 & + 3.98 & \bf 60.21 & + 5.59 & \bf 60.42 & + 5.54 \\
    \midrule
    MoCo-v3 \cite{chen2021mocov3} & 200 & \checkmark & - & 88.89 & - & 60.38 & - & 59.06 & - \\
    \rowcolor{gg}
    \bf Res-MoCo (ours) & 200 & \checkmark & \checkmark & \bf 92.39 & + 3.50 & \bf 64.08 & + 3.70 & \bf 63.32 & + 4.26 \\
    \midrule
    MoCo-v3 \cite{chen2021mocov3} & 400 & \checkmark & - & 89.63 & - & 63.35 & - & 60.64 & - \\
    \rowcolor{gg}
    \bf Res-MoCo (ours) & 400 & \checkmark & \checkmark & \bf 94.43 & + 4.8 & \bf 69.98 & + 6.63 & \bf 67.48 & + 6.84 \\
    \midrule
    MoCo-v3 \cite{chen2021mocov3} & 800 & \checkmark & - & 96.72 & - & 78.14 & - & 72.67 & - \\
    \rowcolor{gg}
    \bf Res-MoCo (ours) & 800 & \checkmark & \checkmark & \bf 98.77 & + 2.05 & \bf 80.41 & + 2.27 & \bf 75.03 & + 2.36 \\
    \midrule
    MoCo-v3 \cite{chen2021mocov3} & 1000 & \checkmark & - & 99.76 & - & 81.26 & - & 75.28 & - \\
    \rowcolor{gg}
    \bf Res-MoCo (ours) & 1000 & \checkmark & \checkmark & \bf 99.94 & + 0.17 & \bf 82.62 & + 1.36 & \bf 75.84 & + 0.56 \\
    \bottomrule
    \end{tabular}}
    \end{center}
    \vspace{-10pt}
    \caption{Similarity of the student and teacher of each image itself on \textbf{ImageNet-100}, \textbf{ResNet-18}. During training, Res-MoCo shows a much higher similarity between teacher and student, which strongly corresponds to a performance boost in both linear top-1 (\%) and KNN accuracy (\%). We monitor similarity (sim) by taking the average cosine similarity of each training batch. }
    \label{tab:CosineSim_imagenet100_res18}
\end{table*}

\subsection{Intra-Momentum for other SSL Frameworks}
\begin{figure}[!htbp]
  \vspace{-10pt}
  \centering
  \includegraphics[width=1.0\linewidth]{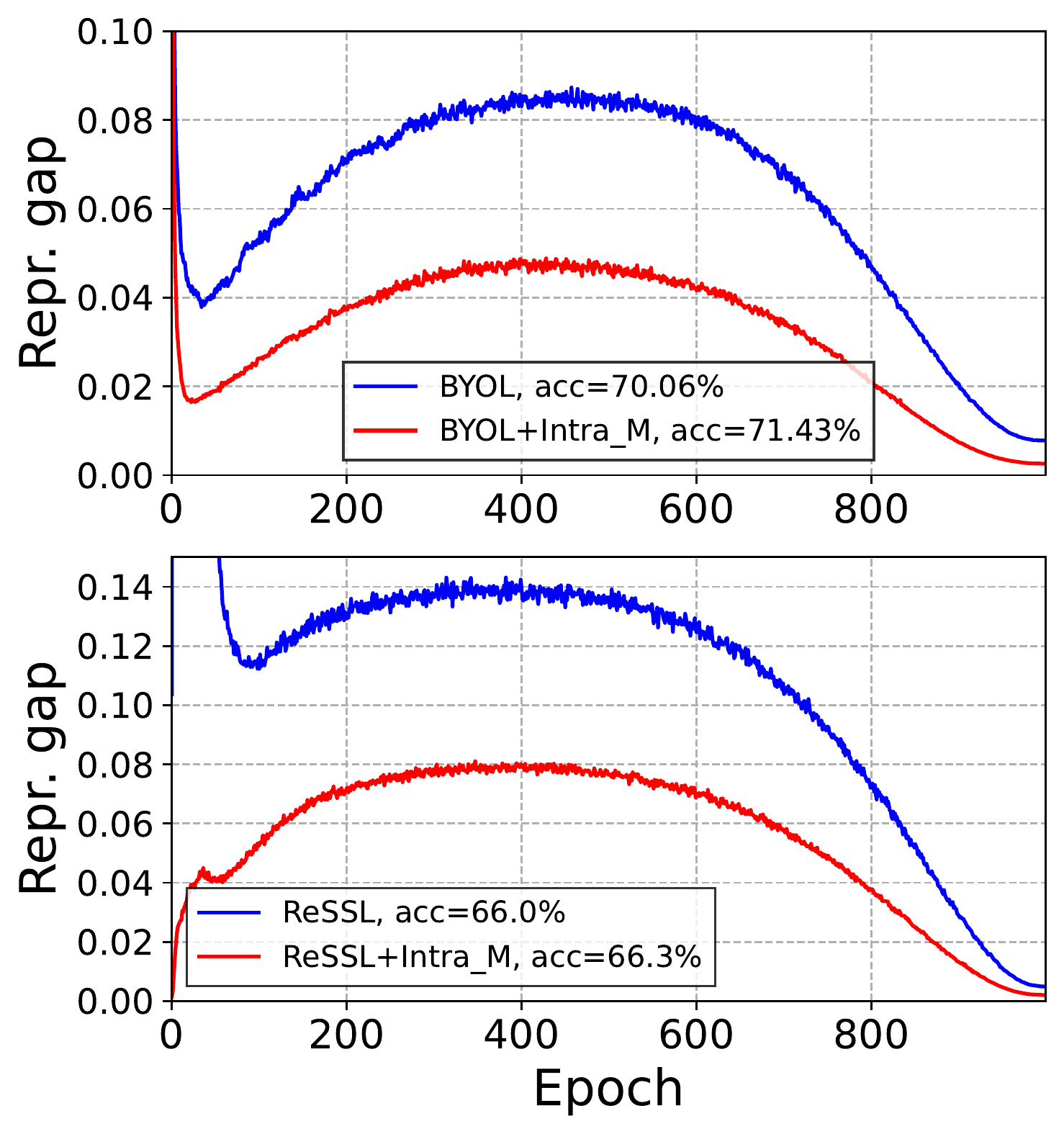}
  \vspace{-16pt}
  \caption{Intra-M with BYOL \cite{grill2020bootstrap} and ReSSL \cite{zheng2021ressl}. All methods are trained for 1000 epochs. It shows a considerable representation gap between the teacher and student for both these EMA-based frameworks.}
  \label{fig:supp_byol_ressl}
  \vspace{-12pt}
\end{figure}

We investigate the usage of the proposed \textit{intra-momentum} in another momentum-free SSL framework such as SimSiam \cite{chen2021exploring}, or the momentum-based ones ReSSL \cite{zheng2021ressl}, BYOL \cite{grill2020bootstrap}. We just inject $\mathcal{L}_{\text{Intra-M}}$ as in Eq. 5 of the manuscript into the loss of those existing SSL frameworks. We report the linear accuracy in Tab. \ref{tab:accuracy_other_ssl} below and Fig. \ref{fig:supp_byol_ressl}. Overall, \textit{intra-momentum} helps to improve both considered momentum-based and momentum-free frameworks. We can also observe that SimSiam framework, the traditional Inter-M does not help, but Intra-M gives a clear improvement from 66.5\% to 67.28\%.
\begin{table}[!htbp]
    \begin{center}
    \resizebox{1.0\hsize}{!}{
    \begin{tabular}{cccccc}
    \toprule
    \bf Method & Inter-M & Intra-M & Top-1 (\%) & Top-5 (\%) & KNN-1 (\%)\\
    \midrule
    \midrule
    BYOL \cite{grill2020bootstrap} & \cmark & - & 70.06 & 92.12 & 61.83 \\
    ours & \cmark & \cmark & \bf 71.43 & \bf 92.54 & \bf 62.34 \\
    \midrule
    ReSSL \cite{zheng2021ressl} & \cmark & - & 66.00 & 89.91 & 59.05 \\
    ours & \cmark & \cmark & \bf 66.30 & \bf 90.17 & \bf 59.32 \\
    \midrule
    SimSiam \cite{chen2021exploring} & - & - & 66.50 & 89.97 & 59.00 \\
     & \cmark & - & 66.33 & 89.71 & 58.59 \\
    ours & - & \cmark & \bf 67.28 & \bf 90.49 \bf & \bf 59.66 \\
    ours & \cmark & \cmark & \bf 67.19 & \bf 90.19 & \bf 59.79 \\
    \bottomrule
    \end{tabular}}
    \end{center}
    \vspace{-14pt}
    \caption{Linear accuracy for other momentum-based and momentum-free SSL frameworks. }
    \label{tab:accuracy_other_ssl}
\end{table}

\subsection{Importance of Intra Momentum in SSL}
\begin{figure}[!htbp]
  \vspace{-10pt}
  \centering
  \includegraphics[width=1.0\linewidth]{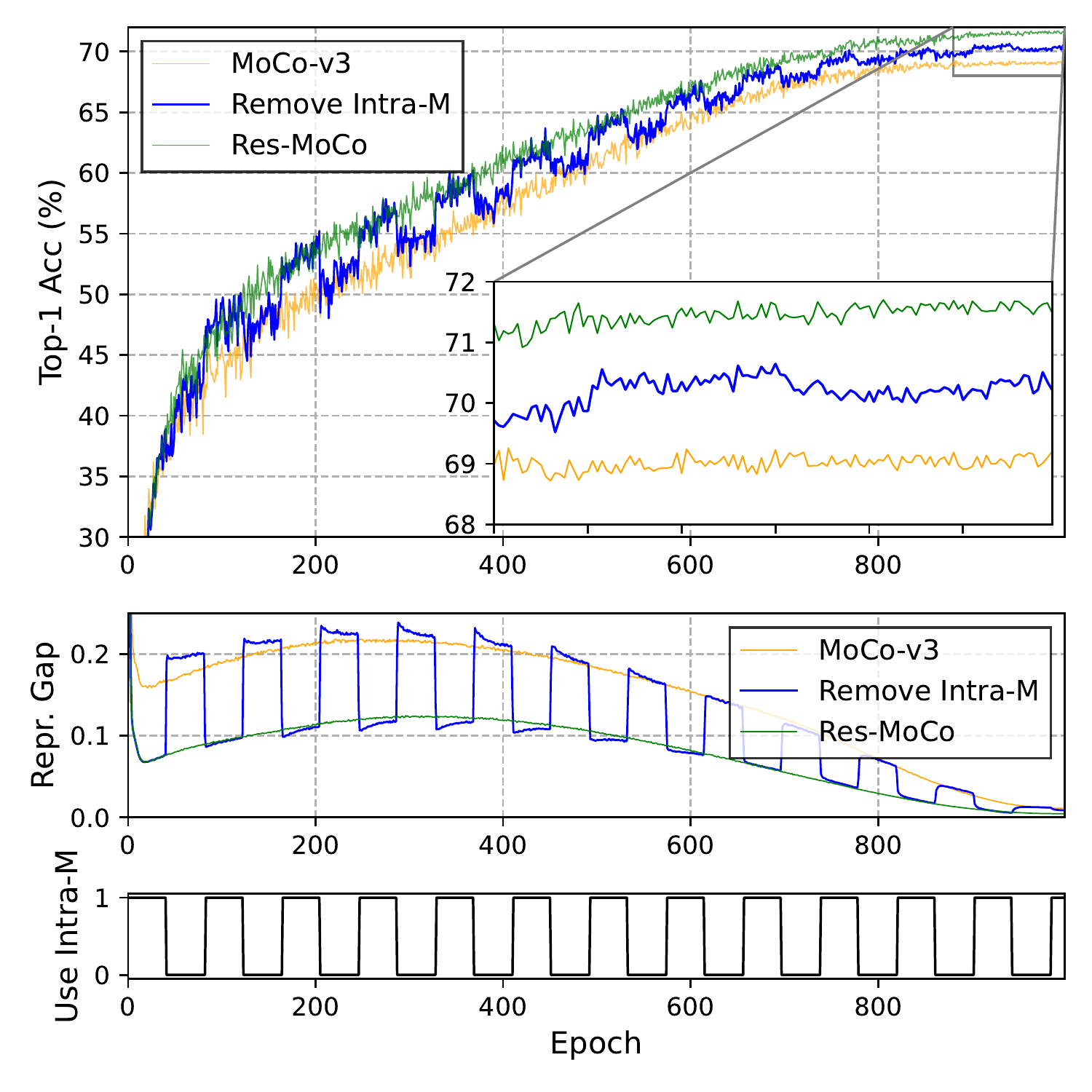}
  \vspace{-16pt}
  \caption{The use of Intra-M, use if value 1 otherwise, not use. We remove Intra-M loss for every 8000 steps ($\sim$ 40 epochs). Experiments were done on CIFAR-100 for 1000 epochs for MoCo-v3 and Res-MoCo. Note that when removing Intra-M, Res-MoCo becomes the baseline MoCo-v3.}
  \label{fig:importance_intraM_all}
  \vspace{-12pt}
\end{figure}

We ablate the importance of the proposed \textit{intra momentum} by injecting \textit{intra momentum} into MoCo-v3 for every 8000 steps during training. We use CIFAR-100 and ResNet-18 to train the SSL model for 1000 epochs when all models fully converge. As shown in Fig. \ref{fig:importance_intraM_all}, whenever adding \textit{intra-M} to the model MoCo-v3, the representation gap (measured by cosine similarity) between the \textit{student} and \textit{teacher} decreases (\textcolor{blue}{blue line}) in Fig. \ref{fig:importance_intraM_all} which strongly corresponds to the performance significantly boost to the level of Res-MoCo (\textcolor{green}{green line}). By contrast, if removing \textit{intra momentum}, we can observe that the accuracy of the student models quickly drops to the MoCo-v3 baseline \textcolor{orange}{orange line}. This indicates the crucial role of the proposed \textit{intra momentum} in the momentum-based SSL framework, i.e. MoCo-v3.

\subsection{Intra Momentum Particularly Benefits for Longer Training}
For supervised learning, training normally needs a short number of epochs, \ie 100 epochs \cite{wang2021solving} to reach a decent performance. By contrast, to get the best performance when the neural networks are fully converged, SSL frameworks need to train for very long epochs, \ie 1000 epochs \cite{grill2020bootstrap,chen2021mocov3,chen2021exploring,zbontar2021barlow}. In this longer training setting with 1000 epochs, momentum-based frameworks often set the momentum coefficient $\beta=0.996$ instead of $\beta=0.99$ for 200 epochs \cite{chen2021exploring,grill2020bootstrap,chen2021mocov3}. As shown in Fig. 8 in the main manuscript, when training 1000 epochs with higher $\beta$ gives better performance; however, the representation gap between teacher and student networks is larger. And the proposed \textit{intra momentum} helps to reduce that gap and significantly improves the performance at long training. In the shorter training, \ie 200 epochs, the $\beta$ is set to 0.99, giving a smaller teacher and student gap, and the models are also not fully converged. This explains why \textit{intra momentum} does not give the best benefit for short training of SSL, but more beneficial for longer training (\ie 1000 epochs) with a higher momentum coefficient (\ie $\beta=0.996$) as common practice \cite{grill2020bootstrap,chen2021mocov3,caron2021emerging,zheng2021ressl}.

\subsection{High Similarity Outputs of the Teacher and Student Gives Better Accuracy}
We transform the representation gap from Eq. 5 in the main manuscript into cosine similarity for a more friendly view when compare their performance on linear evaluation. To this end, we report Tab. \ref{tab:CosineSim_imagenet100_res18} for ResNet-18, and Tab. \ref{tab:CosineSim_imagenet100_res50} for ResNet-50 when training models on ImageNet-100 dataset. It can be seen that over the training, the more similar the representation model has, there better linear accuracy the model obtains. This demonstrates Res-MoCo with the proposed \textit{intra-momentum} highly encourages the student model to match the teacher's output, improving student learning capability.

\subsection{Different Temperature}
We also compare MoCo-v3 \cite{chen2021mocov3} and the proposed Res-MoCo with Intra-M under different temperature parameters. As shown in Tab. \ref{tab:compare_temperature}, Intra-M supports boosting MoCo-v3 for all considered values of temperature $\tau$.

\subsection{More Visualizations}
\textbf{Learned feature maps.} We provide more examples of qualitative results for comparison between MoCo-v3 and the proposed Res-MoCo in the feature maps learning. As shown in Fig. \ref{fig:supp_feature_maps_2}, Res-MoCo shows the learned features with a clearer segmented object than MoCo-v3. We can see the features produced by the proposed Res-MoCo can remove the background noise and focus only on the object compared to MoCo-v3, which remains a lot of redundancies.

\textbf{GradCAM attention map.} We also present more evidence and more diverse examples for attention heatmaps of both methods. Fig. \ref{fig:grad_cam_1} and Fig. \ref{fig:grad_cam_2} reveal more intuitive observations for interpreting where the model looks into the image. It further proves our proposed Res-MoCo successfully better captured the objects in the images compared to MoCo-v3, demonstrating the crucial role of \textit{intra-momentum} in the EMA-based SSL frameworks.

\begin{figure*}[!tbp]
  \centering
  \includegraphics[width=1\linewidth]{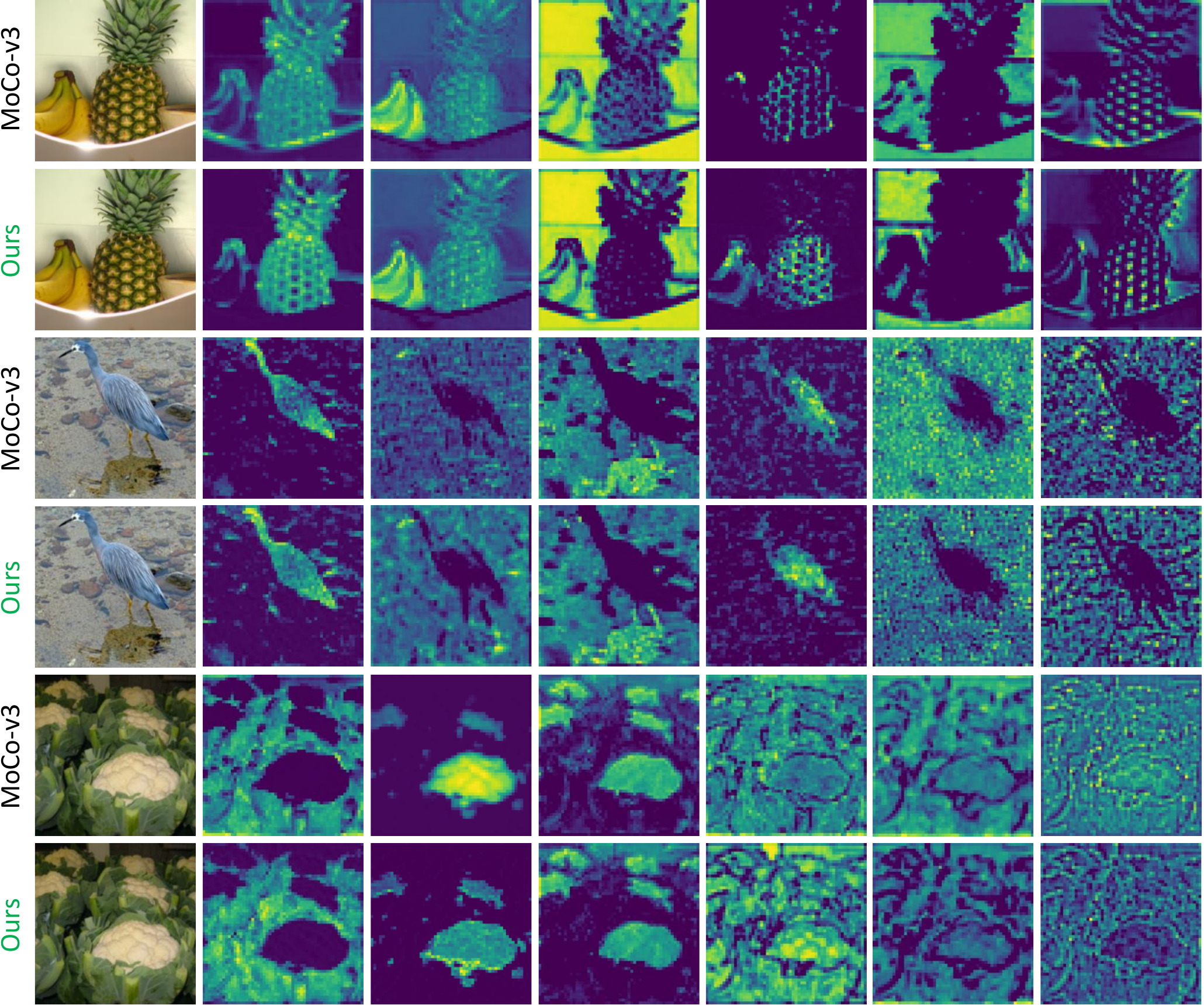}
  \caption{Features learned by MoCo-v3 and Res-MoCo (ours). Some image samples are from the test set of ImageNet-100. We visualize the seven most visually meaningful feature maps of the last convolutional layer in the first block of ResNet-50. It is best viewed in color.}
  \label{fig:supp_feature_maps_1}
\end{figure*}
\begin{figure*}[!tbp]
  \centering
  \includegraphics[width=1\linewidth]{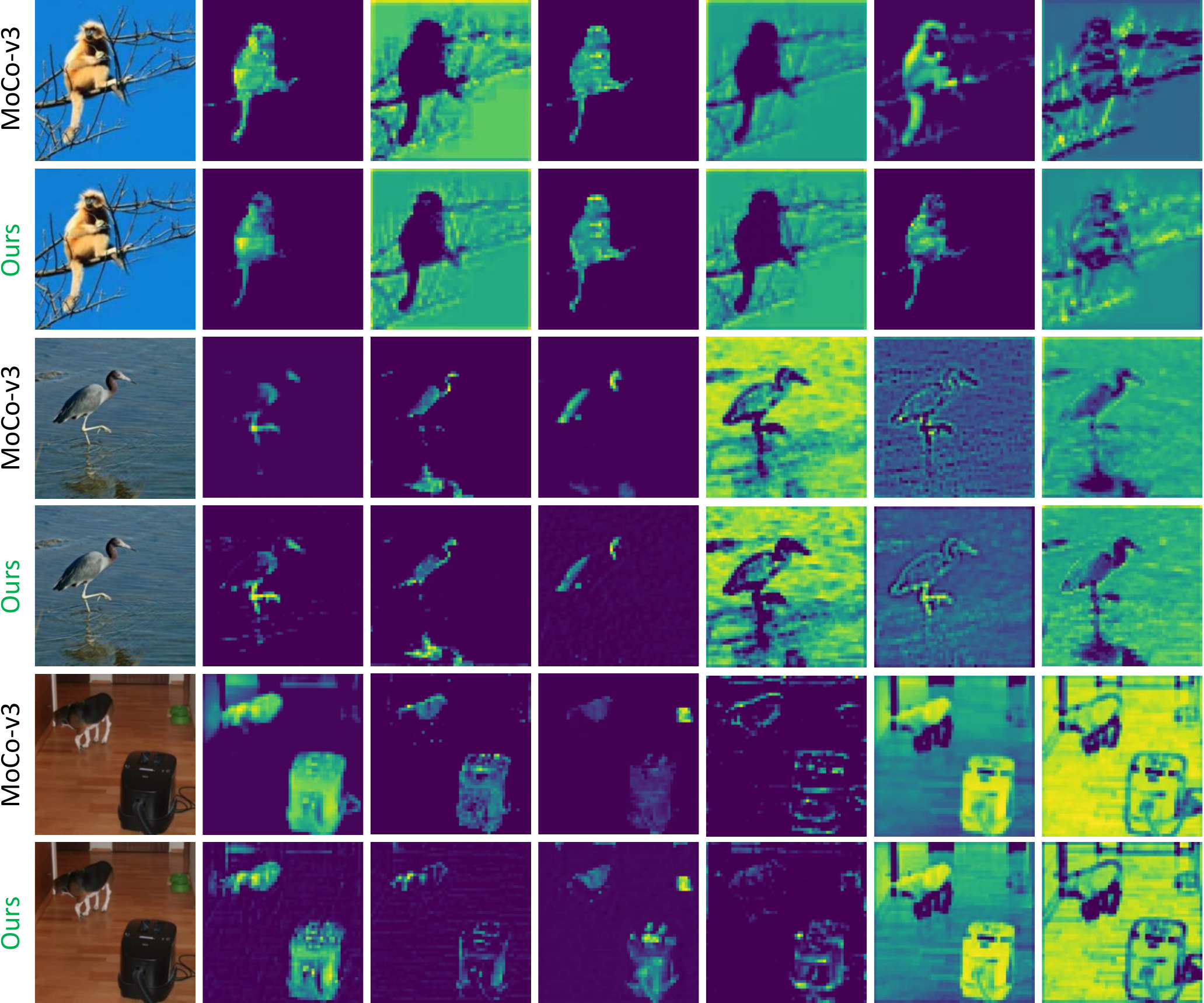}
  \caption{Features learned by MoCo-v3 and Res-MoCo (ours). Some image samples are from the test set of ImageNet-100. We visualize the seven most visually meaningful feature maps of the last convolutional layer in the first block of ResNet-50. It is best viewed in color.}
  \label{fig:supp_feature_maps_2}
\end{figure*}

\begin{figure*}[!tbp]
  \centering
  \includegraphics[width=1.0\linewidth]{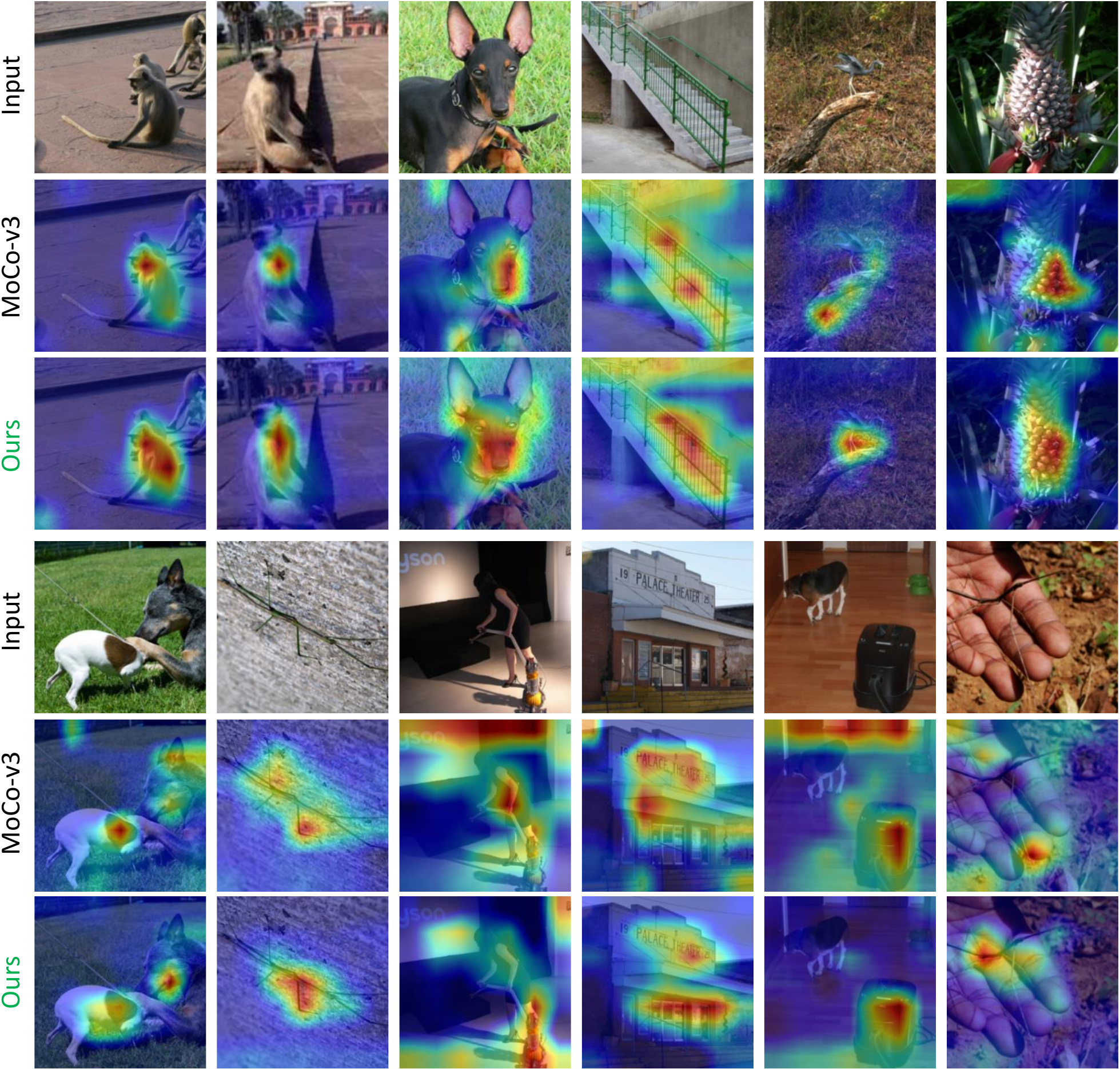}
  \caption{Comparison of GradCAM learned by MoCo-v3 and Res-MoCo (ours). The image samples are from the test set of ImageNet-100. The first row is the original image, the second row is the heat map by MoCo-v3, and the third row is the heat map produced by Res-MoCo (ours). It clearly shows that the heat map produced by Res-MoCo is much more accurate than that of MoCo-v3. It is best viewed in color.}
  \label{fig:grad_cam_1}
\end{figure*}

\begin{figure*}[!tbp]
  \centering
  \includegraphics[width=1.0\linewidth]{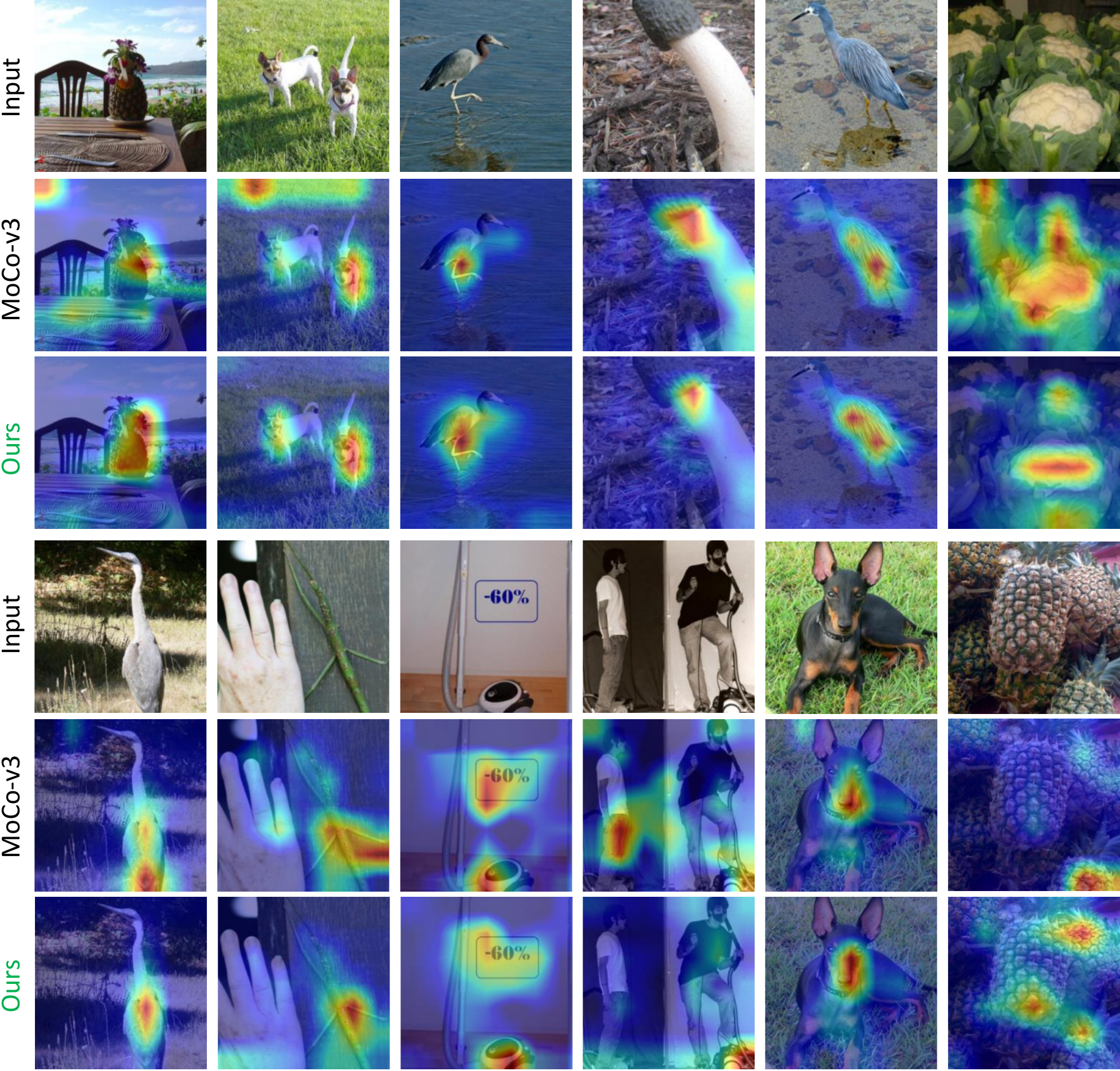}
  \caption{Comparison of GradCAM learned by MoCo-v3 and Res-MoCo (ours). The image samples are from the test set of ImageNet-100. The first row is the original image, the second row is the heat map by MoCo-v3, and the third row is the heat map produced by Res-MoCo (ours). It clearly shows that the heat map produced by Res-MoCo is much more accurate than that of MoCo-v3. It is best viewed in color.}
  \label{fig:grad_cam_2}
\end{figure*}

\end{document}